%% file: main.tex
\documentclass{article}
\usepackage{spconf,amsmath,graphicx}

\input{preamble}

\graphicspath{ {Images/} {Images/Supplementary/} }

\usepackage{subfiles} %

\newcommand{\beginsupplement}{%
    \setcounter{section}{0}
    \setcounter{figure}{0}
    \setcounter{table}{0}
    \setcounter{equation}{0}
    
    \renewcommand{\thesection}{\Alph{section}}
    \renewcommand{\thesubsection}{\thesection.\arabic{subsection}}
    \renewcommand{\thefigure}{S\arabic{figure}}
    \renewcommand{\thetable}{S\arabic{table}}
    \renewcommand{\theequation}{S\arabic{equation}}
    
    \renewcommand{\theHsection}{S\Alph{section}}
    \renewcommand{\theHsubsection}{S\Alph{section}.\arabic{subsection}}
    \renewcommand{\theHfigure}{S\arabic{figure}}
    \renewcommand{\theHtable}{S\arabic{table}}
    \renewcommand{\theHequation}{S\arabic{equation}}
}

\def\x{{\mathbf x}}

\newcommand{\HFS}{H_{\text{FS}}^{K}}
\newcommand{\Hinv}{H^{(-1)}}

\newcommand{\Q}{\mathcal{Q}}

\title{Dithering Defense: Adversarial Robustness of Vision Foundation Models via Multi-Level Floyd--Steinberg Dithering}

\name{Yury Belousov, Brian Pulfer, Vitaliy Kinakh and Slava Voloshynovskiy}
\address{Department of Computer Science, University of Geneva, Switzerland}

\begin{document}
\maketitle
\begin{abstract}
Vision foundation models are widely used as frozen backbones across many downstream tasks, making them a single point of failure under adversarial attack. We study multi-level Floyd--Steinberg error-diffusion dithering as a lightweight, model-agnostic input transformation that disrupts adversarial perturbations while preserving semantic content. Unlike prior work, which was limited to binary dithering, grayscale CIFAR-10, and a single small model trained from scratch, we evaluate across six tasks (classification, segmentation, depth estimation, retrieval, captioning, visual question answering), two model families (DINOv2, PaliGemma), and three attacks of increasing strength (PGD, MI-FGSM, SIA), as well as an adaptive attacker using a straight-through estimator. Our results show that Floyd--Steinberg dithering at intermediate quantization levels, especially when combined with post-processing blur, exceeds or matches all tested baselines, including diffusion-based denoising, with substantially less degradation on clean inputs.
\end{abstract}
\section{Introduction}
\label{sec:introduction}
Vision foundation models (VFMs) have become the default building blocks in computer vision. A single VFM, trained once on a large corpus, can serve as a backbone for classification, segmentation, depth estimation, retrieval, and more, with little or no task-specific fine-tuning. VFMs can also be plugged in as the visual encoder of a vision-language model (VLM), where fine-tuning the backbone risks breaking the learned vision-language alignment. Yet this versatility comes at a cost: because the same frozen backbone is reused everywhere, an adversarial perturbation that fools the backbone compromises every downstream task at once ~\cite{pulfer2025task}. Developing lightweight, model-agnostic defenses that can be applied to already-deployed VFMs, without retraining, is therefore an important and practical problem. Among the possible defenses, input transformations are particularly attractive: they sit in front of the model, require no access to its weights, and can be swapped or combined freely. Classical techniques such as JPEG compression, bit-depth reduction, and blurring have been studied in this role, as well as neural-network-based methods such as diffusion-based denoising; however, each comes with its own trade-offs between clean-image fidelity and adversarial robustness.

Floyd--Steinberg error-diffusion dithering, a classical halftoning technique designed to produce only binary images, has been explored as one such transformation~\cite{lo2021error, ge2020monet}, but prior work leaves major questions open. Evaluation was limited to CIFAR-10 classification ($32{\times}32$ images, 10 classes) with a single small model (ResNet-18) trained from scratch on dithered images, a luxury unavailable for frozen VFMs. The entire study was conducted in grayscale, and the attack repertoire was narrow, excluding transformation-aware attacks designed to survive input pre-processing. Most importantly, only binary (two-level) dithering was tested, leaving the entire spectrum of multi-level quantization unexplored.

We close these gaps with the first comprehensive study of multi-level Floyd--Steinberg dithering as a defense for frozen VFMs. Our evaluation spans six downstream tasks (classification, segmentation, depth estimation, retrieval, visual question answering (VQA), and image captioning), two model families (DINOv2 and PaliGemma), three attacks of increasing strength, including the transformation-robust SIA, across quantization levels from $2$ to $20$. We further stress-test the defense against an adaptive attacker who differentiates through the dithering operation via a straight-through estimator, confirming its robustness even under worst-case conditions.

\begin{figure*}[!t]
    \centering
    \includegraphics[width=0.97\textwidth]{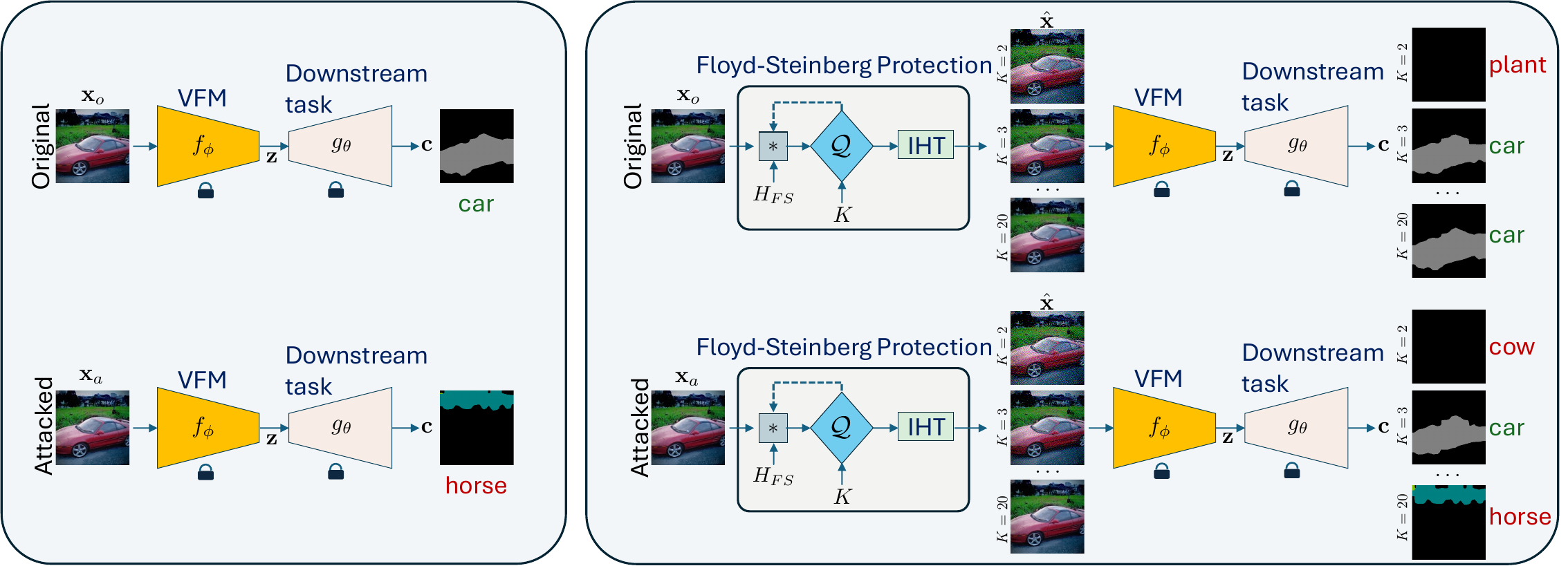}
    \caption{Adversarial defense via Floyd--Steinberg dithering. \textbf{Left:} without defense, the adversarial image $\mathbf{x}_a$ is misclassified and produces a corrupted segmentation mask. \textbf{Right:} inserting FS dithering before the VFM at different quantization levels $K$. Too few levels ($K{=}2$) destroy semantic content; too many ($K{=}20$) let the adversarial perturbation survive; an intermediate value ($K{=}3$) preserves clean accuracy while correctly neutralizing the attack.}
    \label{fig:scheme}
\end{figure*}

\section{Background}
\label{sec:background}

\subsection{Vision foundation models}

VFMs are trained once on large-scale corpora via self-supervised learning, producing general-purpose visual representations that transfer to classification, segmentation, depth estimation, and retrieval with minimal or no fine-tuning.

\subsection{Adversarial attacks}

Adversarial attacks~\cite{szegedy2013intriguing,goodfellow2014explaining} craft small, human-imperceptible perturbations that cause drastic mispredictions. Among them, Projected Gradient Descent (PGD)~\cite{madry18pgd} iteratively maximizes the loss within an $\ell_\infty$-ball, producing strong white-box perturbations. The Momentum Iterative FGSM (MI-FGSM)~\cite{dong2018mifgsm} accumulates gradient momentum across iterations, stabilizing the update direction and improving transferability to unseen models. The Structure Invariant Attack (SIA)~\cite{wang2023sia} goes a step further: it creates multiple copies of the input and applies block-wise random transformations (such as resizing, flipping, rotating, adding noise) to each, then averages the resulting gradients. This produces perturbations that are explicitly optimized to survive input pre-processing, making SIA particularly challenging for transformation-based defenses.

\subsection{Adversarial robustness}
Defenses against adversarial attacks fall into two broad categories. \emph{Adversarial training}~\cite{madry18pgd} augments the training set with adversarial examples, encouraging robust representations. However, it is computationally prohibitive for large-scale pre-training and fundamentally incompatible with the frozen-backbone deployment model of modern foundation models.

\emph{Input transformations}~\cite{guo2017countering} take a complementary, model-agnostic approach: the image is processed before it reaches the model, aiming to destroy adversarial structure while preserving semantic content. A rich set of such transforms has been studied, from classical signal processing (JPEG compression, bit-depth reduction~\cite{xu2017feature}, Gaussian blurring) to neural-network-based approaches such as denoising diffusion smoothing~\cite{carlini2023free, belousov2025beyond}, which passes the input through a pre-trained unconditional diffusion model to remove adversarial perturbations. In the context of input-transformation-based adversarial defense, Floyd--Steinberg error-diffusion dithering has also been investigated~\cite{lo2021error, ge2020monet}.

\section{Dithering Defense}
\label{sec:defense}

Consider a VFM $f_{\phi}$ followed by a task head $g_{\theta}$, producing a task label $\c_o = g_{\theta}(f_{\phi}(\x_o))$ for a clean input $\x_o \in \mathbb{R}^{H \times W \times 3}$ (\cref{fig:scheme}, left). An adversary seeks to find $\x_a$ within an $\epsilon$-ball of $\x_o$ that maximizes a task loss $\mathcal{L}_{c}$, e.g.\ $\mathcal{L}_{c} = \max_{\c\neq \c_o} p(\c|\x) - p(\c_o|\x)$ for classification. We assume a white-box adversary with full access to $f_\phi$ and $g_\theta$, representing a strictly stronger threat than a black-box attack.

The defender seeks to counter this by inserting a pre-processing step $\mathcal{T}$ so that the model sees $\hat{\x} = \mathcal{T}(\x)$ instead of $\x$ (\cref{fig:scheme}, right). For $\mathcal{T}$, we adopt Floyd--Steinberg (FS) error-diffusion dithering~\cite{floyd1976adaptive}, generalized from its original binary setting to $K$ discrete levels per color channel. FS processes the image in raster-scan order: each pixel is quantized to the nearest of $K$ levels by quantizer $\Q$, and the rounding error is distributed to not-yet-visited neighbors via a fixed diffusion kernel. This non-linear, spatially varying, input-dependent transformation is particularly disruptive to the locally coherent perturbation patterns that gradient-based attacks rely on, as we verify in \cref{sec:results}.

We denote the full FS operator as $\HFS(\cdot)$, where $K$ controls quantizer $\Q$, followed by an optional inverse halftone step $\Hinv$ (\cref{fig:scheme}). For $\Hinv$ we use Gaussian blur ($\sigma{=}3$, kernel size $9{\times}9$), which smooths out the high-frequency dithering patterns that can degrade clean performance at low $K$.

\section{Experimental setup}
\label{sec:setup}

To facilitate future research, we release a highly customizable codebase at \url{https://github.com/bruce-willis/attacking-downstream}.

\subsection{Models}
Our primary backbone is DINOv2 ViT-S/14~\cite{oquab2023dinov2}; results with larger variants are reported in the supplementary material (\cref{sec:model_sizes}). We use the task-specific heads released with the model where available and train lightweight linear heads for the remaining tasks.

To test generalization to the multi-modal setting, we also evaluate PaliGemma 3B-mix-224~\cite{beyer2024paligemmaversatile3bvlm}, a vision-language model that pairs a frozen SigLIP~\cite{Zhai_2023_ICCV} visual encoder with a Gemma language decoder. Because the SigLIP backbone remains frozen inside PaliGemma, adversarial training cannot be applied to it without retraining the entire VLM, precisely the scenario motivating this work. 

\subsection{Datasets and metrics}
We evaluate robustness across the following benchmarks and corresponding tasks:
\begin{inparaenum}[(i)]
    \item PascalVOC~\cite{pascal-voc-2012}: \textit{Accuracy} for classification, \textit{Mean Intersection over Union (mIoU)} for segmentation;
    \item NYU-Depth~\cite{couprie2013indoor}: \textit{RMSE} for depth estimation;
    \item Revisited Oxford~\cite{RITAC18}: \textit{Mean Average Precision (mAP)} for image retrieval (easy queries);
    \item COCOCaptions~\cite{chen2015microsoft}: \textit{SPICE}~\cite{anderson2016spice} and \textit{RefCLIPScore}~\cite{hessel2021clipscore} for image captioning;
    \item VQAv2~\cite{balanced_vqa_v2}: \textit{Accuracy} for visual question answering.
\end{inparaenum}

\subsection{Adversarial attacks}

All three attacks (PGD, MI-FGSM, SIA) share the same hyper-parameters: perturbation budget $\epsilon_{\infty} = \frac{3}{255}$ (yielding adversarial images at roughly 40\,dB PSNR), $T = 50$ attack iterations, and step size $\nu = \frac{4\epsilon_{\infty}}{T}$. The attack maximizes a task-specific loss $\mathcal{L}_{c}$: cross-entropy for classification and segmentation, RMSE for depth estimation, and negative cosine similarity between clean and adversarial feature embeddings, $-\cos(f_{\phi}(\x_o), f_{\phi}(\x_a))$, for retrieval. For captioning and VQA we follow~\cite{pulfer2025task} and attack the visual encoder in a task-agnostic manner by maximizing the same negative cosine similarity at the vision-tower output, and then measure the downstream impact on the generated text.

We additionally consider an \emph{informed} (adaptive) threat model in which the attacker knows that the input will be dithered. Because Floyd--Steinberg dithering is non-dif\-fer\-en\-ti\-able (the error-diffusion scan is sequential and input-dependent), we approximate its gradient with a straight-through estimator (STE)~\cite{bengio2013estimating, athalye2018obfuscated}, which passes gradients through the quantization step unchanged. At each attack iteration, the input is optionally quantized to $K_{\text{attack}}$ levels with probability~$p_{\text{q}}$ before being fed into the original attack pipeline. 
Here $K_{\text{attack}}$ plays the same role as the defense-side $K$ but is applied during attack optimization rather than at inference time.
We exhaustively evaluate all combinations of $K_{\text{attack}} \in \{3, 6, 9, 12, 15\}$ and $p_{\text{q}} \in \{0.5, 1.0\}$, yielding quantization-aware variants of all three attacks, and report the worst case.

\section{Results and Discussion}
\label{sec:results}

\cref{fig:scheme} illustrates the central trade-off: too few quantization levels ($K{=}2$) destroy semantic content along with the perturbation; too many ($K{=}20$) let the adversarial signal survive; an intermediate value ($K{=}3$) preserves clean accuracy and neutralizes the attack. Additional qualitative examples are provided in the supplementary material (\cref{sec:qualitative}).

\input{Tables/TableDINOv2Combined}
\input{Tables/TablePaliGemmaCombined}

\subsection{Comparison with other input transformations}
We compare Floyd--Steinberg dithering with a broad set of input transformation baselines described in \cref{sec:background}. The full results are reported in \cref{tab:dinov2_input_transformations}. The model achieves excellent clean performance (94.89\% classification accuracy, 79.26\% segmentation mIoU, 84.46\% retrieval mAP), but is completely vulnerable to adversarial attack: classification accuracy drops to near 0\% under PGD and MI-FGSM, and retrieval mAP collapses to below 1\%.

Among the baselines, none except high-noise diffusion denoised smoothing provides adequate protection under the strongest attack (SIA): JPEG compression, Wiener filtering, geometric transforms, and low-noise diffusion all fall below 33\% classification accuracy. High-noise diffusion is the sole baseline that maintains near-constant performance across clean and attacked inputs, but this comes at the price of severe degradation on clean images relative to the undefended model: classification drops by 14.3 percentage points (pp) (from 94.89\% to 80.60\%), segmentation by 18.0\,pp (79.26\% to 61.30\%), depth RMSE increases from 0.55 to 0.79, and retrieval mAP falls by 22.4\,pp (84.46\% to 62.03\%), an unacceptable trade-off in practice.

Grayscale conversion alone preserves clean performance but provides no adversarial protection. Combining it with quantization or dithering additionally destroys clean performance, as the resulting images deviate too far from the model's training distribution, and re-training foundation models to accommodate such inputs is not practical.

In contrast, color Floyd--Steinberg dithering at $K=3$--$5$, especially when combined with post-processing blur, strikes a notably better balance. FS-based configurations achieve the best or second-best results in 15 out of 16 task--attack combinations (the sole exception being retrieval on clean images, where HFlip and JPEG narrowly lead)\footnote{For retrieval specifically, tuning the blur parameters (rather than using $\sigma{=}3$ across the board) can surpass even these baselines: FS $K{=}3$ with a separately tuned blur reaches 85.7\% clean mAP and 77.3\% under SIA on rOxford, but we keep a single configuration across all tasks for consistency.}. For instance, FS $K=3$ + blur under SIA retains 79.55\% classification accuracy and 64.58\% segmentation mIoU, surpassing even high-noise diffusion (79.42\% and 61.51\%) while maintaining substantially better clean performance (90.83\% vs.\ 80.60\% classification, 72.12\% vs.\ 61.30\% segmentation).

\subsection{Comparison with regular quantization}
We next compare Floyd--Steinberg error diffusion with naive uniform quantization at the same number of levels $K$, to isolate the contribution of the dithering mechanism itself. At $K=3$, regular quantization achieves 85.85\% clean classification accuracy, 80.08\% under PGD, 77.72\% under MI-FGSM, and 65.66\% under SIA, whereas FS dithering reaches 91.74\%, 87.81\%, 85.19\%, and 75.23\%, respectively, a consistent gain of 6--10 percentage points across clean and all attacked conditions. Segmentation follows a similar pattern: regular quantization attains 64.08\% clean mIoU, while FS $K=3$ reaches 72.97\%, and the gap persists under all three attacks. The error diffusion mechanism not only produces more visually faithful images (hence better clean performance), but also distributes adversarial perturbation energy across the image in a way that can be effectively smoothed out, whereas regular quantization simply truncates pixel values and preserves more of the locally coherent adversarial signal.

\subsection{Post-processing via blurring}
Gaussian blurring after Floyd--Steinberg dithering suppresses the high-frequency dithering patterns that degrade clean performance at low $K$, while additionally smoothing residual adversarial structure. At low quantization levels, the effect is dramatic: FS $K=2$ alone yields only 27.52\% clean classification accuracy, but adding blur lifts it to 89.25\%, with SIA robustness improving from 27.79\% to 80.21\%, the best among all methods.
At higher $K$ values, FS dithering without blur leaves the adversarial signal largely intact (e.g., FS $K=8$: 6.42\% under SIA), and adding blur recovers to 
the level of blur applied in isolation.
The sweet spot lies at intermediate levels ($K=3$--$5$), where quantization is coarse enough to genuinely disrupt adversarial perturbations yet fine enough to preserve task-relevant detail. 

\subsection{Informed (adaptive) attack}
All attacks considered so far are \emph{oblivious} to the defense. We now evaluate the stronger adaptive threat model, focusing on informed SIA, the strongest oblivious attack by a wide margin. \cref{tab:informed_attack} compares the oblivious SIA result (O) with the worst case over all $(K_{\text{attack}}, p_{\text{q}})$ combinations (I).

The key finding is that FS + blur configurations are remarkably resilient: the worst-case informed SIA degrades classification by only 2.2\,pp for FS $K{=}3$ + blur (79.55\% $\to$ 77.33\%) and 2.1\,pp for FS $K{=}4$ + blur (77.72\%~$\to$~75.62\%). Segmentation follows the same pattern, with drops of at most 1.4\,pp (mIoU). Depth RMSE and retrieval mAP are similarly stable for FS + blur. In contrast, FS \emph{without} blur is more vulnerable: FS $K{=}4$ alone drops by 9.6\,pp under informed SIA. This confirms that Gaussian blur acts as a second line of defense, smoothing any residual adversarial signal that survives the STE-based attack. Compared with other defenses under the same adaptive threat, FS $K{=}3$ + blur outperforms high-noise diffusion in segmentation (63.15\% vs.\ 59.79\% mIoU) and Gaussian blur (drops 2.4\,pp %
for classification), confirming the value of combining quantization with smoothing.

\input{Tables/TableInformedCompact}

\looseness=-1
We further analyze $K_{\text{attack}}$ vs.\ $p_{\text{q}}$ in the supplement (\cref{sec:informed_details}). Notably, $p_{\text{q}}{=}1.0$ weakens the attack: constant quantization 
over-constrains the STE gradients, so that even on the \emph{undefended} model, accuracy jumps from 0.39\% to 91.61\%.

\subsection{Vision-language model}
The results in \cref{tab:paligemma_input_transformations} confirm that the defense generalizes to multi-modal PaliGemma: intermediate quantization levels again offer the best robustness--fidelity trade-off, consistent with the DINOv2 experiments. This demonstrates that the defense is model-agnostic and its benefits extend beyond classification and segmentation to generative language outputs conditioned on visual features.

\section{Conclusion}
\label{sec:conclusion}

We presented a comprehensive study of multi-level Floyd--Steinberg error-diffusion dithering as an input-transformation defense for vision foundation models. Across six downstream tasks, two model families, and three attacks of increasing strength, FS dithering at intermediate quantization levels, 
combined with post-processing blur, consistently exceeded or matched all tested baselines, including diffusion-based denoising, with substantially less degradation on clean inputs. The defense remained effective even under an adaptive attacker equipped with a straight-through estimator, with worst-case degradation of only 2\,pp for FS $K{=}3$ + blur, and generalized from a pure vision backbone (DINOv2) to a vision-language model (PaliGemma) without modification.

Several directions remain open. The current scheme applies the same number of quantization levels to all color channels; per-channel levels could exploit the varying sensitivity of different color channels to adversarial perturbations. Operating in alternative color spaces (e.g., YCbCr) before quantization, using non-uniform level spacing, or introducing controlled randomness into the dithering process are further avenues that may improve the robustness--fidelity trade-off.

\vfill\pagebreak

\clearpage

{
\ninept
\bibliographystyle{IEEEbib}
\bibliography{refs}
}

\clearpage
\beginsupplement

\subfile{supplement}

\end{document}

%% file: preamble.tex
\usepackage{comment} %
\usepackage{amssymb} %
\usepackage{etoolbox} %
\usepackage{pdflscape} %
\usepackage[linesnumbered,ruled,vlined]{algorithm2e} %

\SetCommentSty{commentfont}
\usepackage{url} %
\usepackage{xr-hyper} %
\usepackage[pagebackref,breaklinks]{hyperref} %
\usepackage[capitalize]{cleveref} %

\usepackage[caption=false]{subfig} %

\usepackage{paralist} %

\usepackage[dvipsnames]{xcolor} %

\usepackage{adjustbox} %

\usepackage{booktabs} %
\usepackage{multirow} %
\usepackage{multicol} %
\usepackage{longtable} %
\newcommand{\itseries}{\fontshape{it}\selectfont} %
\robustify\itseries
\newrobustcmd{\IT}{\itseries} 
\usepackage{arydshln} %

\usepackage{pifont} %

\newcommand{\x}{\mathbf{x}}

\renewcommand{\c}{\mathbf{c}}

%% file: Tables/TableDINOv2Combined.tex
\begin{table*}[htbp]
\caption{Clean and adversarial performance of DINOv2 across four downstream tasks under different input transformations. FS denotes Floyd--Steinberg dithering with $K$ quantization levels per channel; blur denotes Gaussian blur ($\sigma=3$, kernel size $9$). Best and second-best attacked results are in \textbf{bold} and \underline{underlined}. $\uparrow$: higher is better; $\downarrow$: lower is better.}
\label{tab:dinov2_input_transformations}
\centering
\resizebox{\linewidth}{!}{
\begin{tabular}{lcccc|cccc|cccc|cccc}
\toprule
\multirow{4}{*}{Method} & \multicolumn{8}{c}{Pascal VOC} & \multicolumn{4}{c}{NYUD v2} & \multicolumn{4}{c}{ROxford5k} \\
\cmidrule(lr){2-9} \cmidrule(lr){10-13} \cmidrule(lr){14-17}
 & \multicolumn{4}{c}{Classification $\uparrow$} & \multicolumn{4}{c}{Segmentation $\uparrow$} & \multicolumn{4}{c}{Depth Error $\downarrow$} & \multicolumn{4}{c}{mAP $\uparrow$} \\
\cmidrule(lr){2-5} \cmidrule(lr){6-9} \cmidrule(lr){10-13} \cmidrule(lr){14-17}
 & Clean & PGD & MI-FGSM & SIA & Clean & PGD & MI-FGSM & SIA & Clean & PGD & MI-FGSM & SIA & Clean & PGD & MI-FGSM & SIA \\
\midrule
 \IT None & \IT 94.89 & \IT 0.00 & \IT 0.00 & \IT 0.39 & \IT 79.26 & \IT 10.50 & \IT 13.48 & \IT 21.26 & \IT 0.55 & \IT 6.41 & \IT 6.28 & \IT 4.11 & \IT 84.46 & \IT 0.46 & \IT 0.47 & \IT 0.63 \\
 HFlip & 94.76 & 31.45 & 6.29 & 1.18 & 79.07 & 46.80 & 27.62 & 23.38 & 0.56 & 2.12 & 2.93 & 3.91 & \textbf{85.49} & 82.02 & 40.47 & 4.02 \\
 VFlip & 83.36 & 38.27 & 25.16 & 1.18 & 57.75 & 44.70 & 38.90 & 23.92 & 0.95 & 1.56 & 1.75 & 3.32 & 46.24 & 19.43 & 10.52 & 1.16 \\
 Blur & 90.83 & 88.60 & 86.24 & 69.72 & 71.16 & 69.42 & 68.06 & 57.53 & 0.68 & 0.74 & 0.78 & 0.92 & 82.08 & 81.63 & 80.08 & 75.08 \\
 Wiener & 92.92 & 74.31 & 58.85 & 32.50 & 71.68 & 62.84 & 53.67 & 37.85 & 0.67 & 0.84 & 0.98 & 1.39 & 82.24 & 77.49 & 71.67 & 53.82 \\
 JPEG & 94.63 & 52.69 & 16.38 & 2.75 & 78.56 & 57.83 & 39.39 & 25.63 & 0.57 & 1.41 & 2.21 & 3.62 & \underline{84.62} & \textbf{82.72} & 56.81 & 12.84 \\
 Grayscale & 94.36 & 6.55 & 1.31 & 1.97 & 78.71 & 24.72 & 19.13 & 23.54 & 0.59 & 5.03 & 5.31 & 4.01 & 83.84 & 28.99 & 5.39 & 4.09 \\
 Rotation & 82.57 & 49.15 & 36.30 & 3.93 & 59.49 & 49.84 & 45.66 & 26.32 & 1.16 & 1.23 & 1.26 & 2.40 & 28.55 & 20.94 & 16.59 & 3.69 \\
 Diffusion (low) & \textbf{95.54} & 46.26 & 9.17 & 2.23 & 79.03 & 54.56 & 34.63 & 25.98 & 0.58 & 1.42 & 2.52 & 3.66 & 83.82 & 81.66 & 50.07 & 15.42 \\
 Diffusion (high) & 80.60 & 79.42 & 79.03 & 79.42 & 61.30 & 61.45 & 61.32 & 61.51 & 0.79 & 0.79 & 0.78 & 0.81 & 62.03 & 64.55 & 66.18 & 58.21 \\
 Quant. Gray & 30.93 & 31.45 & 31.32 & 30.41 & 31.79 & 32.18 & 32.39 & 31.95 & 1.53 & 1.52 & 1.51 & 1.53 & 5.98 & 6.24 & 6.22 & 5.82 \\
 Quant. Gray + blur & 50.59 & 49.54 & 49.80 & 47.44 & 41.85 & 41.66 & 41.82 & 41.41 & 1.33 & 1.32 & 1.31 & 1.32 & 41.01 & 41.98 & 41.79 & 40.90 \\
 FS Gray & 5.11 & 5.37 & 5.37 & 5.24 & 26.30 & 26.01 & 26.19 & 25.90 & 1.85 & 1.83 & 1.84 & 1.84 & 0.74 & 0.75 & 0.74 & 0.78 \\
 FS Gray + blur & 79.42 & 79.42 & 78.90 & 75.75 & 60.06 & 59.80 & 59.52 & 58.24 & 0.74 & 0.75 & 0.76 & 0.79 & 58.15 & 58.01 & 57.83 & 57.92 \\
 \midrule
 Quant. $K=2$ & 61.86 & 59.50 & 58.06 & 54.65 & 43.95 & 43.31 & 43.49 & 41.42 & 1.22 & 1.23 & 1.24 & 1.28 & 42.92 & 39.42 & 38.71 & 37.07 \\
 Quant. $K=2$ + blur & 63.96 & 64.22 & 64.61 & 62.52 & 45.73 & 46.20 & 45.79 & 44.67 & 1.14 & 1.13 & 1.12 & 1.13 & 51.68 & 52.51 & 51.92 & 51.45 \\
 FS $K=2$ & 27.52 & 29.23 & 29.49 & 27.79 & 35.71 & 36.14 & 35.82 & 35.58 & 1.76 & 1.77 & 1.77 & 1.77 & 2.50 & 2.11 & 2.38 & 2.09 \\
 FS $K=2$ + blur & 89.25 & 88.34 & 87.42 & \textbf{80.21} & 70.53 & 69.41 & 68.47 & \underline{64.55} & 0.67 & 0.70 & \underline{0.71} & \textbf{0.76} & 72.35 & 73.14 & 73.30 & 69.75 \\
 Quant. $K=3$ & 85.85 & 80.08 & 77.72 & 65.66 & 64.08 & 61.50 & 60.04 & 51.96 & 0.74 & 0.78 & 0.80 & 0.97 & 77.65 & 76.03 & 74.61 & 70.47 \\
 Quant. $K=3$ + blur & 76.41 & 75.49 & 74.97 & 73.39 & 52.23 & 52.42 & 53.16 & 50.99 & 0.85 & 0.85 & 0.85 & 0.88 & 66.00 & 64.69 & 65.06 & 64.44 \\
 FS $K=3$ & 91.74 & 87.81 & 85.19 & 75.23 & 72.97 & 71.03 & 69.04 & 60.83 & 0.59 & \textbf{0.64} & \textbf{0.67} & 0.87 & 77.12 & 75.29 & 76.31 & 72.03 \\
 FS $K=3$ + blur & 90.83 & 88.99 & \textbf{88.07} & \underline{79.55} & 72.12 & 71.42 & \textbf{70.25} & \textbf{64.58} & 0.68 & 0.71 & 0.73 & \underline{0.80} & 77.14 & 75.65 & 75.07 & 71.66 \\
 FS $K=4$ & 93.32 & 84.80 & 75.75 & 55.44 & 76.19 & \textbf{71.79} & 65.35 & 50.21 & 0.55 & \underline{0.65} & 0.78 & 1.24 & 83.43 & 82.08 & 79.53 & 71.15 \\
 FS $K=4$ + blur & 90.83 & \textbf{89.52} & \underline{87.94} & 77.72 & 72.72 & \underline{71.53} & \underline{70.24} & 62.91 & 0.68 & 0.73 & 0.75 & 0.82 & 82.56 & 79.19 & 75.03 & 71.78 \\
 FS $K=5$ & 94.23 & 78.37 & 62.12 & 33.29 & 77.20 & 69.22 & 58.99 & 41.20 & 0.54 & 0.72 & 0.94 & 1.72 & 84.37 & \underline{82.22} & 75.19 & 61.13 \\
 FS $K=5$ + blur & 91.09 & \underline{89.12} & 87.16 & 75.49 & 72.44 & 71.19 & 69.83 & 61.78 & 0.68 & 0.73 & 0.77 & 0.85 & 80.79 & 79.79 & 79.09 & 72.33 \\
 FS $K=6$ & 93.84 & 71.30 & 48.23 & 18.35 & 78.00 & 65.86 & 52.17 & 34.99 & 0.53 & 0.80 & 1.11 & 2.20 & 84.24 & 81.69 & 71.99 & 47.43 \\
 FS $K=6$ + blur & 91.22 & 88.86 & 87.68 & 73.39 & 72.59 & 70.73 & 69.27 & 60.67 & 0.68 & 0.74 & 0.77 & 0.87 & 81.89 & 81.41 & 80.04 & 73.26 \\
 FS $K=8$ & 93.71 & 56.49 & 24.90 & 6.42 & 78.26 & 58.64 & 41.78 & 28.87 & \underline{0.52} & 0.98 & 1.52 & 2.90 & 84.08 & 80.59 & 57.13 & 27.02 \\
 FS $K=8$ + blur & 90.83 & 88.47 & 87.42 & 72.08 & 72.18 & 70.37 & 69.04 & 59.41 & 0.68 & 0.74 & 0.78 & 0.89 & 82.06 & 81.20 & \textbf{80.72} & 74.30 \\
 FS $K=10$ & 94.76 & 41.68 & 10.88 & 2.10 & \underline{79.16} & 51.85 & 34.04 & 26.35 & \textbf{0.51} & 1.18 & 1.95 & 3.29 & 84.49 & 76.42 & 39.08 & 10.50 \\
 FS $K=10$ + blur & 90.83 & 88.47 & 86.63 & 70.64 & 71.92 & 70.03 & 68.59 & 58.71 & 0.68 & 0.74 & 0.78 & 0.90 & 82.14 & 81.47 & \underline{80.58} & \textbf{75.94} \\
 FS $K=15$ & 94.76 & 20.45 & 1.31 & 0.79 & 78.75 & 38.25 & 23.00 & 23.30 & \textbf{0.51} & 1.79 & 3.17 & 3.74 & 84.40 & 56.74 & 7.83 & 2.84 \\
 FS $K=15$ + blur & 90.96 & 88.73 & 86.63 & 70.12 & 71.45 & 69.63 & 68.29 & 58.13 & 0.68 & 0.74 & 0.78 & 0.91 & 82.09 & 81.80 & 80.16 & \underline{75.37} \\
 FS $K=20$ & \underline{94.89} & 7.86 & 0.52 & 0.66 & \textbf{79.20} & 28.47 & 17.61 & 22.41 & \underline{0.52} & 2.56 & 4.37 & 3.90 & 84.31 & 28.26 & 3.87 & 1.60 \\
 FS $K=20$ + blur & 90.83 & 88.34 & 86.37 & 70.38 & 71.29 & 69.56 & 68.14 & 57.81 & 0.68 & 0.74 & 0.78 & 0.92 & 81.98 & 81.88 & 80.11 & 74.80 \\
\bottomrule
\end{tabular}
}
\end{table*} 

%% file: Tables/TablePaliGemmaCombined.tex
\begin{table*}[htbp]
\caption{Clean and adversarial performance of PaliGemma on COCOCaptions and VQAv2 under different input transformations. FS denotes Floyd--Steinberg dithering with $K$ quantization levels per channel; blur denotes Gaussian blur ($\sigma=3$, kernel size $9$). Best and second-best attacked results are in \textbf{bold} and \underline{underlined}. $\uparrow$: higher is better.}
\label{tab:paligemma_input_transformations}
\centering
\resizebox{\linewidth}{!}{
\begin{tabular}{lcccc|cccc|cccc|cccc|cccc}
\toprule
\multirow{4}{*}{Method} & \multicolumn{8}{c}{CocoCaptions} & \multicolumn{12}{c}{VQAv2} \\
\cmidrule(lr){2-9} \cmidrule(lr){10-21}
 & \multicolumn{4}{c}{SPICE $\uparrow$} & \multicolumn{4}{c}{RefCLIPScore $\uparrow$} & \multicolumn{4}{c}{yes/no $\uparrow$} & \multicolumn{4}{c}{number $\uparrow$} & \multicolumn{4}{c}{other $\uparrow$} \\
\cmidrule(lr){2-5} \cmidrule(lr){6-9} \cmidrule(lr){10-13} \cmidrule(lr){14-17} \cmidrule(lr){18-21}
 & Clean & PGD & MI-FGSM & SIA & Clean & PGD & MI-FGSM & SIA & Clean & PGD & MI-FGSM & SIA & Clean & PGD & MI-FGSM & SIA & Clean & PGD & MI-FGSM & SIA \\
\midrule
 \IT None & \IT 24.85 & \IT 12.24 & \IT 11.91 & \IT 5.79 & \IT 83.50 & \IT 69.25 & \IT 67.80 & \IT 55.62 & \IT 95.94 & \IT 82.44 & \IT 81.87 & \IT 70.97 & \IT 72.69 & \IT 50.29 & \IT 48.80 & \IT 32.44 & \IT 77.12 & \IT 51.79 & \IT 50.27 & \IT 32.70 \\
 Blur & 18.75 & 18.48 & 18.41 & 16.53 & 77.56 & 77.56 & 77.56 & 75.56 & 90.98 & 90.54 & 90.44 & 87.92 & 56.89 & 56.37 & 56.26 & 52.08 & 66.88 & 66.32 & 65.92 & 61.48 \\
 Grayscale & \underline{22.59} & 19.37 & 16.98 & 9.09 & \textbf{82.00} & \textbf{79.00} & 76.00 & 63.90 & 93.10 & 88.95 & 85.88 & 75.47 & 68.03 & 61.43 & 57.70 & 42.11 & 64.41 & 56.84 & 51.79 & 35.64 \\
 Diffusion (high) & 15.17 & 15.17 & 15.11 & 15.00 & 74.06 & 74.00 & 74.00 & 73.75 & 82.50 & 82.49 & 82.50 & 82.25 & 43.13 & 42.86 & 43.06 & 42.70 & 50.88 & 50.54 & 50.74 & 50.39 \\
\midrule
 FS $K=2$ & 17.17 & 17.03 & 17.03 & 15.85 & 74.75 & 74.60 & 74.56 & 73.06 & 87.49 & 87.28 & 87.35 & 85.66 & 52.83 & 52.77 & 52.68 & 50.46 & 60.57 & 60.42 & 60.28 & 57.47 \\
 FS $K=2$ + blur & 18.78 & 18.55 & 18.47 & \textbf{16.79} & 78.40 & \underline{78.25} & \textbf{78.10} & \textbf{76.44} & 90.78 & 90.52 & 90.42 & \textbf{88.33} & 56.79 & 56.43 & 56.22 & 52.73 & 66.99 & 66.48 & 66.17 & \textbf{62.37} \\
 FS $K=3$ & 19.26 & 19.00 & 18.75 & \underline{16.75} & 76.94 & 76.70 & 76.44 & 74.00 & 92.29 & 91.89 & 91.48 & \textbf{88.33} & 62.88 & 61.69 & 61.29 & \textbf{55.90} & 69.22 & 68.47 & 67.79 & \underline{62.15} \\
 FS $K=3$ + blur & 18.73 & 18.43 & 18.35 & 16.59 & 78.00 & 77.80 & \underline{77.80} & \underline{76.00} & 90.85 & 90.46 & 90.36 & \underline{88.03} & 56.78 & 56.31 & 56.04 & 52.11 & 66.88 & 66.26 & 65.90 & 61.67 \\
 FS $K=4$ & 19.75 & 19.40 & \textbf{18.97} & 15.95 & 77.40 & 77.00 & 76.56 & 72.75 & 93.62 & 93.03 & \underline{92.34} & 87.57 & 65.97 & 64.70 & 63.67 & \underline{55.66} & 72.06 & 70.69 & \underline{69.56} & 60.96 \\
 FS $K=4$ + blur & 18.68 & 18.44 & 18.43 & 16.54 & 77.80 & 77.80 & 77.69 & 75.90 & 90.85 & 90.57 & 91.86 & 87.93 & 56.74 & 56.20 & 63.13 & 52.20 & 66.90 & 66.30 & 68.64 & 61.70 \\
 FS $K=5$ & 19.99 & \underline{19.54} & \underline{18.84} & 14.45 & 77.50 & 77.10 & 76.30 & 70.44 & 94.21 & 93.38 & \textbf{92.44} & 85.47 & 67.78 & 65.96 & \textbf{64.12} & 53.60 & 73.35 & 71.60 & \textbf{69.60} & 57.63 \\
 FS $K=6$ & 20.22 & \textbf{19.61} & 18.52 & 12.72 & 77.60 & 77.10 & 75.90 & 67.56 & 94.64 & \textbf{93.62} & 92.02 & 82.74 & 69.10 & \textbf{66.21} & \underline{63.71} & 50.44 & 74.19 & \underline{71.82} & 68.83 & 53.13 \\
 FS $K=7$ & 20.42 & 19.53 & 18.09 & 11.13 & 77.80 & 77.00 & 75.30 & 64.90 & 94.89 & \underline{93.54} & 91.48 & 80.37 & 69.97 & \underline{66.20} & 62.88 & 47.30 & 74.69 & \textbf{71.87} & 67.72 & 49.06 \\
 FS $K=8$ & 20.62 & 19.47 & 17.54 & 9.90 & 78.00 & 76.90 & 74.70 & 62.60 & 95.15 & 93.40 & 90.70 & 78.21 & 70.47 & 65.88 & 61.92 & 44.32 & 75.19 & 71.45 & 66.08 & 45.48 \\
 FS $K=10$ & 21.05 & 19.09 & 16.48 & 8.45 & 78.40 & 76.44 & 73.25 & 59.80 & 95.43 & 92.91 & 89.13 & 75.67 & 71.15 & 65.01 & 59.71 & 39.89 & 75.56 & 70.40 & 63.03 & 40.88 \\
 FS $K=12$ & 21.57 & 18.64 & 15.56 & 7.66 & 78.90 & 76.00 & 72.00 & 58.30 & 95.53 & 92.18 & 87.70 & 74.21 & 71.70 & 63.96 & 57.59 & 37.59 & 75.90 & 68.92 & 60.34 & 38.40 \\
 FS $K=15$ & 22.54 & 17.94 & 14.56 & 7.02 & \underline{80.06} & 75.30 & 70.75 & 57.10 & \underline{95.67} & 90.96 & 86.05 & 73.09 & \underline{72.11} & 62.25 & 55.07 & 35.71 & \underline{76.21} & 66.64 & 57.44 & 36.35 \\
 FS $K=20$ & \textbf{24.00} & 16.81 & 13.51 & 6.48 & \textbf{82.00} & 74.25 & 69.70 & 56.34 & \textbf{95.85} & 89.06 & 84.32 & 72.21 & \textbf{72.34} & 59.51 & 52.83 & 34.22 & \textbf{76.72} & 63.06 & 54.54 & 34.76 \\
\bottomrule
\end{tabular}
}
\end{table*}

%% file: Tables/TableInformedCompact.tex
\begin{table}[tb]
\caption{Performance under oblivious (O) vs.\ worst-case informed (I) SIA attack. $\Delta$: degradation from adaptive attacker. FS + blur degrades by only 1--3\,pp across all tasks.}
\label{tab:informed_attack}
\centering
\resizebox{\columnwidth}{!}{
\begin{tabular}{l ccc ccc cc cc}
\toprule
\multirow{2}{*}{Method} & \multicolumn{3}{c}{Cls.\ Acc.\ $\uparrow$} & \multicolumn{3}{c}{Seg.\ mIoU $\uparrow$} & \multicolumn{2}{c}{Depth $\downarrow$} & \multicolumn{2}{c}{mAP $\uparrow$} \\
\cmidrule(lr){2-4} \cmidrule(lr){5-7} \cmidrule(lr){8-9} \cmidrule(lr){10-11}
 & O & I & $\Delta$ & O & I & $\Delta$ & O & I & O & I \\
\midrule
 \IT Blur & \IT 69.72 & \IT 67.37 & \IT $-$2.4 & \IT 57.53 & \IT 56.59 & \IT $-$0.9 & \IT 0.92 & \IT 0.94 & \IT 75.08 & \IT 73.23 \\
\midrule
 FS $K\!=\!2$ + blur & 80.21 & 79.42 & $-$0.8 & 64.55 & 62.85 & $-$1.7 & 0.76 & 0.77 & 69.75 & 69.22 \\
 FS $K\!=\!3$ & 75.23 & 69.33 & $-$5.9 & 60.83 & 57.23 & $-$3.6 & 0.87 & 0.96 & 72.03 & 67.59 \\
 FS $K\!=\!3$ + blur & 79.55 & 77.33 & $-$2.2 & 64.58 & 63.15 & $-$1.4 & 0.80 & 0.79 & 71.66 & 70.83 \\
 FS $K\!=\!4$ & 55.44 & 45.87 & $-$9.6 & 50.21 & 45.38 & $-$4.8 & 1.24 & 1.40 & 71.15 & 64.00 \\
 FS $K\!=\!4$ + blur & 77.72 & 75.62 & $-$2.1 & 62.91 & 61.65 & $-$1.3 & 0.82 & 0.82 & 71.78 & 70.80 \\
 FS $K\!=\!5$ & 33.29 & 25.95 & $-$7.3 & 41.20 & 37.61 & $-$3.6 & 1.72 & 1.96 & 61.13 & 54.73 \\
 FS $K\!=\!5$ + blur & 75.49 & 72.74 & $-$2.8 & 61.78 & 60.47 & $-$1.3 & 0.85 & 0.85 & 72.33 & 71.67 \\
\bottomrule
\end{tabular}
}
\end{table}

%% file: supplement.tex
\makeatletter
\@ifclassloaded{subfiles}{%
    \externaldocument{main} 
    \beginsupplement
    \twocolumn[{%
        \centering
        \vspace{1em}
        {\Large \textbf{Dithering Defense: Adversarial Robustness of Vision Foundation Models via Multi-Level Floyd--Steinberg Dithering}}\\[0.5em] %
        {\large Supplementary Material}\\[1.5em]
    }]
}{}
\makeatother

\section{Qualitative examples: the role of quantization granularity}
\label{sec:qualitative}

\nameCref{tab:qualitative_examples}~\hyperlink{hyper:tab:qualitative}{\ref*{tab:qualitative_examples}} presents a per-image walkthrough that lays bare how the quantization level $K$ shapes the defense on both clean and adversarial inputs. Six PascalVOC examples are passed through Floyd--Steinberg dithering at $K \in \{2, 3, 6, 9, 20\}$ and fed into DINOv2 ViT-S/14 for classification and segmentation; ground-truth labels, predicted classes with confidence scores, and predicted segmentation masks are displayed side by side.

The two extremes bracket the design space. At $K{=}2$ the image is quantized so aggressively that even clean inputs are stripped of recognizable structure: every example collapses to a spurious category (e.g., ``pottedplant,'' ``bottle''), and the segmentation masks bear little resemblance to the actual scene content. At the other end, $K{=}20$ is effectively transparent to the adversarial perturbation. The classification confidences under attack at $K{=}20$ are nearly indistinguishable from those of the undefended model: the targeted adversarial classes typically retain their exact peak confidences, with the largest deviation across the entire set being a mere $\sim$9\% drop (as seen in example~\hyperlink{hyper:qual:a}{(a)}). This confirms that such a fine quantization grid offers no meaningful bottleneck against gradient-based perturbations.

The most revealing behavior lies between these poles. At $K{=}3$ (the configuration our quantitative results in \cref{tab:dinov2_input_transformations} identify as the sweet spot), the defense succeeds uniformly: every adversarial image is classified correctly with high confidence ($>83\%$ across all six examples), and the segmentation maps closely track the ground truth. This level sits in the narrow band where quantization is coarse enough to shatter the locally coherent adversarial signal yet fine enough to preserve the semantic features the backbone relies on.

At $K{=}6$, the defense enters a \emph{transitional regime} that is particularly instructive. Rather than a clean binary outcome (correct or fooled), several examples exhibit partial penetration of the adversarial signal: the adversarial target class now dominates the softmax, yet the true class has not vanished entirely. In example~\hyperlink{hyper:qual:c}{(c)}, ``bottle'' reaches 68.6\% while the ground-truth ``person'' still registers at 28.3\%; in example~\hyperlink{hyper:qual:d}{(d)}, ``sheep'' sits at 59.9\% with ``cow'' trailing at 18.5\%; and in example~\hyperlink{hyper:qual:a}{(a)}, ``motorbike'' leads at 72.0\% but ``car'' persists at 8.2\%. The segmentation maps at this level frequently maintain recognizable object boundaries (the car silhouette in~\hyperlink{hyper:qual:a}{(a)}, the dog outline in~\hyperlink{hyper:qual:b}{(b)}) even though the classifier has already tipped toward the adversarial label. This dual presence of the true and adversarial classes signals that the quantization bottleneck at $K{=}6$ partially, but not sufficiently, disrupts the perturbation: enough adversarial structure survives to sway the softmax, yet the semantic content of the original scene has not been fully overwritten. Notably, example~\hyperlink{hyper:qual:f}{(f)} remains correctly classified as ``aeroplane'' at 98.4\% confidence even at $K{=}6$ (and 91.8\% at $K{=}9$), illustrating that the precise threshold at which the defense breaks down varies with image content and the geometry of the perturbation. As $K$ grows beyond this transitional zone, the residual true-class probability drops steadily until it vanishes, and predictions converge to the undefended adversarial output observed at $K{=}20$.

\section{Scaling to Larger Backbones}
\label{sec:model_sizes}

The main paper (\cref{sec:setup}) reports all results using DINOv2 ViT-S/14 (21\,M parameters). We now evaluate the defense on two larger variants within the same model family: ViT-B/14 (86\,M) and ViT-L/14 (300\,M), focusing on the image retrieval task on rOxford5k (easy queries). The same experimental protocol (\cref{sec:setup}) is used: PGD, MI-FGSM, and SIA attacks with $\epsilon_\infty = 3/255$, and Floyd--Steinberg dithering with optional Gaussian blur ($\sigma{=}3$, kernel size $9{\times}9$).

\textbf{\emph{Vulnerability is universal.}}
Without any defense, all three backbones are equally catastrophically vulnerable: mAP collapses to below 1.6\% under every attack, regardless of model size (\cref{tab:model_size_comparison}, first row). This confirms that larger capacity alone offers no protection against adversarial perturbations.

\begin{figure*}[htbp]
    \centering
    \includegraphics[width=\linewidth]{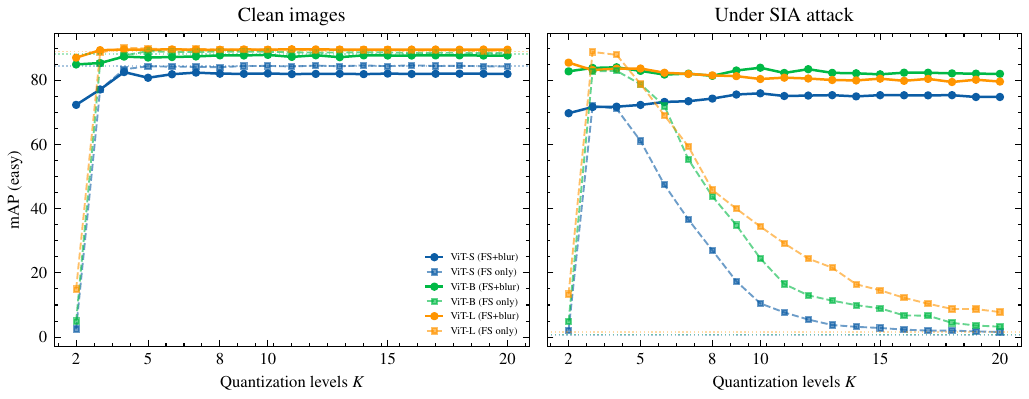}
    \caption{Retrieval mAP (easy) on rOxford5k as a function of quantization levels $K$ for three DINOv2 backbone sizes. Solid lines with circles: FS + blur; dashed lines with squares: FS without blur. Dotted horizontal lines: undefended baselines. \textbf{Left:} clean images. \textbf{Right:} under SIA attack. Larger models tolerate coarser quantization and show less clean degradation.}
    \label{fig:map_vs_k}
\end{figure*}

\textbf{\emph{Defense effectiveness increases with model size.}}
\cref{tab:model_size_comparison} and \cref{fig:map_vs_k} show that the defense becomes more effective as the backbone grows.
Under SIA, the strongest oblivious attack, FS $K{=}3$ without blur recovers mAP from near zero to 72.03\% for ViT-S, 82.76\% for ViT-B, and 88.85\% for ViT-L. The last figure is remarkable: ViT-L with FS $K{=}3$ achieves 88.85\% SIA mAP, nearly matching its own clean baseline of 88.91\%.
Clean performance also degrades less for larger models: FS $K{=}3$ loses 7.3\,pp of clean mAP on ViT-S but only 0.3\,pp on ViT-L; FS $K{=}4$ actually \emph{improves} ViT-L clean mAP by 1.2\,pp (from 88.91\% to 90.15\%).

\textbf{\emph{The role of post-processing blur shifts with model size.}}
On ViT-S, blur is consistently beneficial or neutral: it smooths dithering artifacts that degrade retrieval features. For ViT-L, however, blur hurts SIA robustness at low $K$: FS $K{=}3$ alone achieves 88.85\% SIA mAP, but adding blur drops it to 83.18\% ($-$5.7\,pp). This is visible in the right panel of \cref{fig:map_vs_k}, where the dashed curves (FS only) lie above the solid curves (FS + blur) for ViT-L at low $K$. A plausible explanation is that larger models are more tolerant of dithering artifacts (their representations are already robust to such high-frequency patterns), so the blur removes useful discriminative structure along with the artifacts.

\begin{figure*}[htbp]
    \centering
    \includegraphics[width=0.6\linewidth]{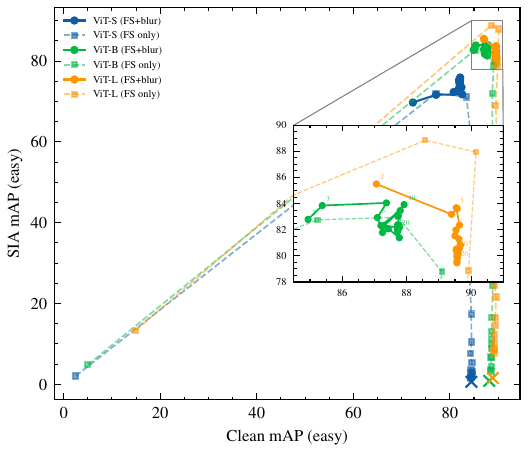}
    \caption{Clean mAP vs.\ SIA mAP trade-off on rOxford5k for Floyd--Steinberg dithering ($K{=}2$--$20$) across three DINOv2 sizes. Solid: FS + blur; dashed: FS only. Crosses ($\times$): no defense (near-zero SIA mAP). The Pareto frontier shifts upward with model size; ViT-L achieves near-baseline clean \emph{and} adversarial mAP simultaneously.}
    \label{fig:tradeoff}
\end{figure*}

\textbf{\emph{Clean vs.\ adversarial trade-off.}}
\cref{fig:tradeoff} plots clean mAP against SIA mAP for all $K$ values and model sizes. The Pareto frontier shifts upward and to the right with model size: ViT-L configurations occupy the top-right corner (high clean and high adversarial mAP), while ViT-S configurations are pushed toward the lower left. The ideal operating point moves from FS $K{=}10$ + blur for ViT-S (82.14\% clean, 75.94\% SIA) to FS $K{=}3$ alone for ViT-L (88.58\% clean, 88.85\% SIA), the latter achieving near-perfect robustness with negligible clean degradation.

\section{Informed attack: detailed analysis}
\label{sec:informed_details}

\begin{figure*}[htbp]
    \centering
    \includegraphics[width=0.85\textwidth]{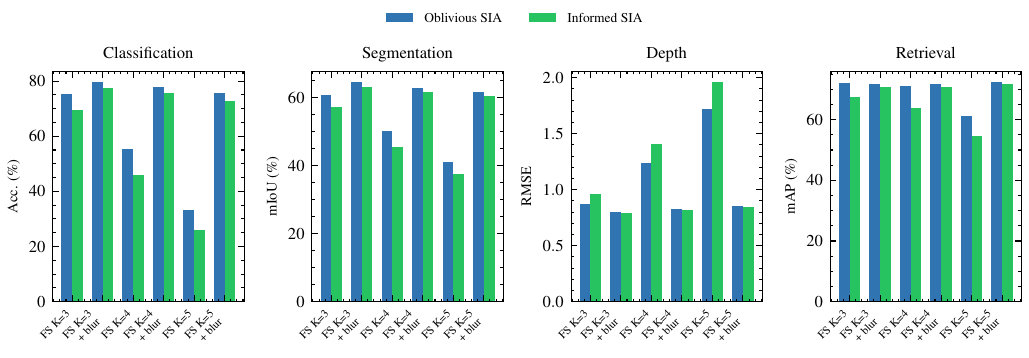}
    \caption{Oblivious vs.\ worst-case informed SIA across four downstream tasks. FS + blur defenses show minimal degradation across all tasks.}
    \label{fig:informed_bar}
\end{figure*}

\begin{figure*}[htbp]
    \centering
    \subfloat[Classification $\uparrow$ (PascalVOC)]{\includegraphics[width=\textwidth]{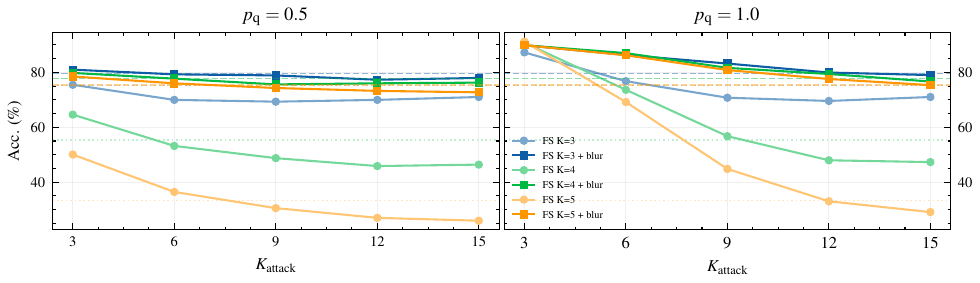}\label{fig:informed_sia_classification}}\\[-0.5ex]
    \subfloat[Segmentation $\uparrow$ (PascalVOC)]{\includegraphics[width=\textwidth]{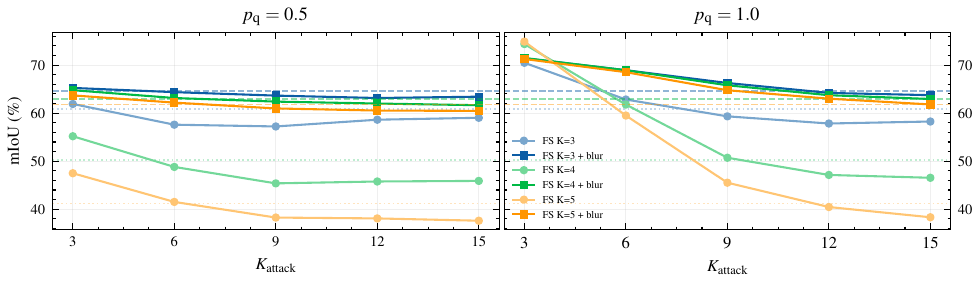}\label{fig:informed_sia_segmentation}}\\[-0.5ex]
    \subfloat[Depth estimation $\downarrow$ (NYU-Depth)]{\includegraphics[width=\textwidth]{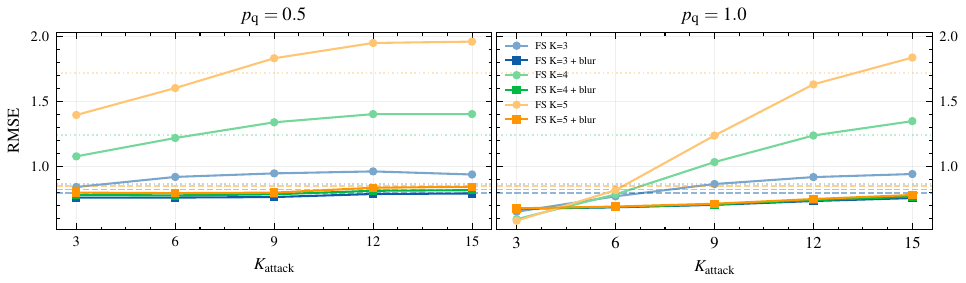}\label{fig:informed_sia_depth}}\\[-0.5ex]
    \subfloat[Retrieval $\uparrow$ (Revisited Oxford)]{\includegraphics[width=\textwidth]{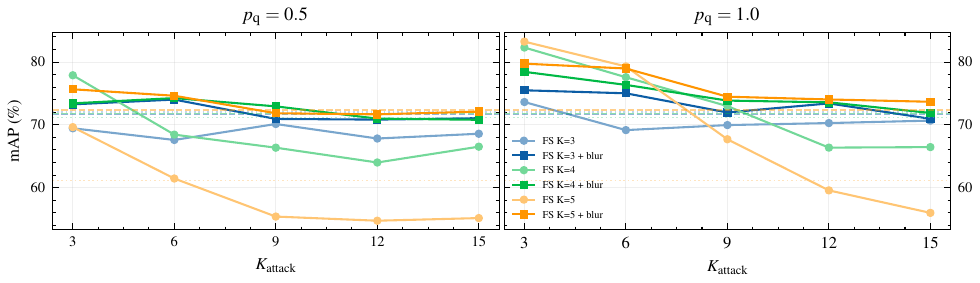}\label{fig:informed_sia_retrieval}}
    \caption{Informed SIA analysis across all four tasks. Each subfigure shows a different task; columns correspond to $p_{\text{q}}{=}0.5$ (left) and $p_{\text{q}}{=}1.0$ (right). Circle markers: FS without blur; square markers: FS + blur. Horizontal baselines show oblivious SIA performance (dashed for +blur, dotted for no-blur). The robustness of FS + blur generalizes consistently across all tasks.}
    \label{fig:informed_sia_all}
\end{figure*}

\cref{fig:informed_sia_all} shows how the best-performing FS configurations ($K{=}3$--$5$) hold up under informed SIA across all four downstream tasks as a function of the attacker's quantization level $K_{\text{attack}}$ for two values of the dithering probability $p_{\text{q}}$. Horizontal lines show oblivious SIA baselines in matching colors: dashed lines for defenses with blur (square markers), dotted lines for defenses without blur (circle markers). The trends are unambiguous and consistent across all tasks:
\begin{inparaenum}[(i)]
    \item FS + blur configurations (square markers) remain nearly flat across all $K_{\text{attack}}$ values, confirming that the post-processing blur absorbs most of the additional adversarial signal the STE enables;
    \item at $p_{\text{q}}{=}1.0$ (right column), performance is generally \emph{higher} than the oblivious baseline, because constant quantization during attack optimization over-constrains the perturbation search;
    \item remarkably, FS $K{=}3$ \emph{without} blur (blue circles) is itself nearly as robust as its blurred counterpart---it tracks closely behind and never suffers the dramatic collapse seen at $K{=}4$ and $K{=}5$ once the attacker increases $K_{\text{attack}}$.
\end{inparaenum}
This last point deserves emphasis: the coarse three-level quantization grid alone provides a strong enough bottleneck to blunt the informed attack, making blur a helpful but not strictly necessary addition at this operating point. The pattern is consistent across classification~\subref{fig:informed_sia_classification}, segmentation~\subref{fig:informed_sia_segmentation}, and retrieval~\subref{fig:informed_sia_retrieval}; depth estimation~\subref{fig:informed_sia_depth} tells the same story for $K{=}3$, though without-blur variants at $K{=}4$ and $K{=}5$ degrade more sharply as $K_{\text{attack}}$ grows, with RMSE nearly tripling in the worst case. \cref{fig:informed_bar} complements this analysis, visualizing the oblivious--informed gap side by side for all four tasks.

To quantify these trends, \cref{tab:kattack_effect} zooms in on FS $K{=}3$ + blur --- the best-performing configuration --- and reports classification accuracy for every $(K_{\text{attack}}, p_{\text{q}})$ pair. The worst case at $p_{\text{q}}{=}0.5$ occurs at $K_{\text{attack}}{=}12$ (77.33\%), still only 2.2\,pp below the oblivious baseline of 79.55\% (see \cref{tab:informed_attack}); at $p_{\text{q}}{=}1.0$, every $K_{\text{attack}}$ value yields accuracy \emph{above} the oblivious baseline, corroborating the over-constraining effect discussed above. \cref{tab:informed_full} extends this analysis to the full cross-product of all four tasks and all three attack types (PGD, MI-FGSM, SIA), confirming that the resilience of FS + blur is not limited to SIA or to classification alone.

\input{Tables/TableKattackEffect}

\input{Tables/TableModelSizeComparison_highlighted}

\input{Tables/TableInformedFull}

{
\clearpage
\onecolumn

\begin{landscape}
    \input{Tables/TableQualitativeExamples}
\end{landscape}
\clearpage

\twocolumn
}

%% file: Tables/TableKattackEffect.tex
\begin{table}[htbp]
\caption{Effect of the attacker's quantization level $K_{\text{attack}}$ and dithering probability $p_{\text{q}}$ on FS $K{=}3$ + blur under informed SIA attack (PascalVOC classification accuracy). The oblivious SIA baseline is 79.55\%. Higher $p_{\text{q}}$ counter-intuitively weakens the attack by over-constraining the STE gradients.}
\label{tab:kattack_effect}
\centering
\begin{tabular}{c cc}
\toprule
$K_{\text{attack}}$ & $p_{\text{q}} = 0.5$ & $p_{\text{q}} = 1.0$ \\
\midrule
3 & 81.00 & 89.91 \\
6 & 79.29 & 86.37 \\
9 & 78.90 & 83.22 \\
12 & 77.33 & 79.95 \\
15 & 77.98 & 79.03 \\
\bottomrule
\end{tabular}
\end{table}

%% file: Tables/TableModelSizeComparison_highlighted.tex
\begin{table*}[htbp]
\caption{Image retrieval (mAP, easy queries) on rOxford5k for three DINOv2 backbone sizes under different FS dithering configurations. All attacks use $\epsilon_\infty = 3/255$. Best and second-best attacked results per model size are in \textbf{bold} and \underline{underlined}. $\uparrow$: higher is better.}
\label{tab:model_size_comparison}
\centering
\resizebox{\linewidth}{!}{
\begin{tabular}{lcccc|cccc|cccc}
\toprule
\multirow{3}{*}{Method} & \multicolumn{4}{c}{ViT-S} & \multicolumn{4}{c}{ViT-B} & \multicolumn{4}{c}{ViT-L} \\
\cmidrule(lr){2-5} \cmidrule(lr){6-9} \cmidrule(lr){10-13} 
 & \multicolumn{4}{c}{mAP $\uparrow$} & \multicolumn{4}{c}{mAP $\uparrow$} & \multicolumn{4}{c}{mAP $\uparrow$} \\
\cmidrule(lr){2-5} \cmidrule(lr){6-9} \cmidrule(lr){10-13} 
 & Clean & PGD & MI-FGSM & SIA & Clean & PGD & MI-FGSM & SIA & Clean & PGD & MI-FGSM & SIA \\
\midrule
 \IT None & \IT 84.46 & \IT 0.46 & \IT 0.47 & \IT 0.63 & \IT 88.17 & \IT 0.46 & \IT 0.50 & \IT 0.87 & \IT 88.91 & \IT 0.46 & \IT 0.65 & \IT 1.56 \\
 FS $K$=2 + blur & 72.35 & 73.14 & 73.30 & 69.75 & 84.94 & 85.05 & 84.91 & 82.79 & 87.06 & 86.55 & 86.94 & 85.50 \\
 FS $K$=3 & 77.12 & 75.29 & 76.31 & 72.03 & 85.23 & 85.08 & 85.99 & 82.76 & 88.58 & 88.86 & \textbf{89.84} & \textbf{88.85} \\
 FS $K$=3 + blur & 77.14 & 75.65 & 75.07 & 71.66 & 85.38 & 85.44 & 85.93 & 83.85 & 89.39 & 89.17 & 89.03 & 83.18 \\
 FS $K$=4 & \underline{83.43} & \underline{82.08} & 79.53 & 71.15 & 87.74 & \textbf{88.93} & 87.22 & 82.95 & \textbf{90.15} & \textbf{89.90} & \underline{89.65} & \underline{87.93} \\
 FS $K$=4 + blur & 82.56 & 79.19 & 75.03 & 71.78 & 87.37 & 86.39 & 87.87 & \textbf{84.05} & 89.57 & 88.66 & 88.99 & 83.62 \\
 FS $K$=5 & \textbf{84.37} & \textbf{82.22} & 75.19 & 61.13 & \textbf{89.09} & \underline{88.61} & 87.16 & 78.80 & \underline{89.92} & \underline{89.19} & 88.21 & 78.88 \\
 FS $K$=5 + blur & 80.79 & 79.79 & 79.09 & 72.33 & 87.09 & 86.02 & 86.17 & 82.93 & 89.54 & 88.71 & 88.77 & 83.68 \\
 FS $K$=8 + blur & 82.06 & 81.20 & \textbf{80.72} & 74.30 & 87.77 & 88.35 & 87.45 & 81.39 & 89.50 & 88.33 & 88.83 & 81.54 \\
 FS $K$=10 + blur & 82.14 & 81.47 & \underline{80.58} & \textbf{75.94} & \underline{87.92} & 88.47 & \textbf{88.83} & \underline{83.92} & 89.54 & 88.46 & 89.13 & 80.40 \\
 FS $K$=15 + blur & 82.09 & 81.80 & 80.16 & \underline{75.37} & 87.72 & 88.28 & \underline{88.66} & 81.86 & 89.54 & 88.95 & 89.01 & 80.51 \\
 FS $K$=20 + blur & 81.98 & 81.88 & 80.11 & 74.80 & 87.73 & 88.23 & 88.65 & 82.00 & 89.56 & 88.48 & 89.11 & 79.60 \\
\bottomrule
\end{tabular}
}
\end{table*}

%% file: Tables/TableInformedFull.tex
\begin{table*}[htbp]
\caption{Detailed comparison: oblivious (O) vs.\ worst-case informed (I) attack across four tasks and three attack types. For each cell, we report the metric under the oblivious attack and the worst case over all $(K_{\text{attack}}, p_{\text{q}})$ combinations of the informed attack. $\uparrow$: higher is better; $\downarrow$: lower is better.}
\label{tab:informed_full}
\centering
\resizebox{\linewidth}{!}{
\begin{tabular}{l cc|cc|cc cc|cc|cc cc|cc|cc cc|cc|cc}
\toprule
\multirow{3}{*}{Method} & \multicolumn{6}{c}{Classification $\uparrow$} & \multicolumn{6}{c}{Segmentation $\uparrow$} & \multicolumn{6}{c}{Depth Error $\downarrow$} & \multicolumn{6}{c}{mAP $\uparrow$} \\
\cmidrule(lr){2-7} \cmidrule(lr){8-13} \cmidrule(lr){14-19} \cmidrule(lr){20-25}
 & \multicolumn{2}{c}{PGD} & \multicolumn{2}{c}{MI-FGSM} & \multicolumn{2}{c}{SIA} & \multicolumn{2}{c}{PGD} & \multicolumn{2}{c}{MI-FGSM} & \multicolumn{2}{c}{SIA} & \multicolumn{2}{c}{PGD} & \multicolumn{2}{c}{MI-FGSM} & \multicolumn{2}{c}{SIA} & \multicolumn{2}{c}{PGD} & \multicolumn{2}{c}{MI-FGSM} & \multicolumn{2}{c}{SIA} \\
\cmidrule(lr){2-3} \cmidrule(lr){4-5} \cmidrule(lr){6-7} \cmidrule(lr){8-9} \cmidrule(lr){10-11} \cmidrule(lr){12-13} \cmidrule(lr){14-15} \cmidrule(lr){16-17} \cmidrule(lr){18-19} \cmidrule(lr){20-21} \cmidrule(lr){22-23} \cmidrule(lr){24-25}
 & O & I & O & I & O & I & O & I & O & I & O & I & O & I & O & I & O & I & O & I & O & I & O & I \\
\midrule
Blur & 88.60 & 87.68 & 86.24 & 83.75 & 69.72 & 67.37 & 69.42 & 69.03 & 68.06 & 66.27 & 57.53 & 56.59 & 0.74 & 0.74 & 0.78 & 0.81 & 0.92 & 0.91 & 81.63 & 79.98 & 80.08 & 78.49 & 75.08 & 72.39 \\
Diffusion ($t=396$) & 79.16 & 78.11 & 80.47 & 77.72 & 79.95 & 77.98 & 61.48 & 59.78 & 61.59 & 60.31 & 62.02 & 59.79 & 0.78 & 0.81 & 0.80 & 0.81 & 0.79 & 0.81 & 62.15 & 60.40 & 63.47 & 58.15 & 62.61 & 60.08 \\
\midrule
FS $K=2$ + blur & 88.34 & 87.55 & 87.42 & 85.98 & 80.21 & 79.42 & 69.41 & 69.15 & 68.47 & 67.79 & 64.55 & 62.85 & 0.70 & 0.70 & 0.71 & 0.73 & 0.76 & 0.77 & 73.14 & 71.66 & 73.30 & 68.82 & 69.75 & 67.74 \\
FS $K=3$ & 87.81 & 84.93 & 85.19 & 77.98 & 75.23 & 69.33 & 71.03 & 69.36 & 69.04 & 64.96 & 60.83 & 57.23 & 0.64 & 0.67 & 0.67 & 0.78 & 0.87 & 0.96 & 75.29 & 73.06 & 76.31 & 69.96 & 72.03 & 67.59 \\
FS $K=3$ + blur & 88.99 & 88.86 & 88.07 & 86.89 & 79.55 & 77.33 & 71.42 & 70.98 & 70.25 & 69.14 & 64.58 & 63.15 & 0.71 & 0.71 & 0.73 & 0.75 & 0.80 & 0.79 & 75.65 & 74.60 & 75.07 & 73.55 & 71.66 & 70.83 \\
FS $K=4$ & 84.80 & 79.16 & 75.75 & 63.30 & 55.44 & 45.87 & 71.79 & 68.78 & 65.35 & 57.13 & 50.21 & 45.38 & 0.65 & 0.70 & 0.78 & 1.01 & 1.24 & 1.40 & 82.08 & 78.36 & 79.53 & 71.15 & 71.15 & 64.00 \\
FS $K=4$ + blur & 89.52 & 88.60 & 87.94 & 86.37 & 77.72 & 75.62 & 71.53 & 71.34 & 70.24 & 68.94 & 62.91 & 61.65 & 0.73 & 0.72 & 0.75 & 0.77 & 0.82 & 0.82 & 79.19 & 77.66 & 75.03 & 75.25 & 71.78 & 70.80 \\
FS $K=5$ & 78.37 & 69.46 & 62.12 & 40.63 & 33.29 & 25.95 & 69.22 & 64.30 & 58.99 & 47.90 & 41.20 & 37.61 & 0.72 & 0.78 & 0.94 & 1.33 & 1.72 & 1.96 & 82.22 & 77.85 & 75.19 & 65.51 & 61.13 & 54.73 \\
FS $K=5$ + blur & 89.12 & 87.94 & 87.16 & 86.24 & 75.49 & 72.74 & 71.19 & 70.84 & 69.83 & 68.57 & 61.78 & 60.47 & 0.73 & 0.73 & 0.77 & 0.79 & 0.85 & 0.85 & 79.79 & 77.64 & 79.09 & 75.98 & 72.33 & 71.67 \\
FS $K=6$ & 71.30 & 60.42 & 48.23 & 23.98 & 18.35 & 12.98 & 65.86 & 59.76 & 52.17 & 41.53 & 34.99 & 33.02 & 0.80 & 0.87 & 1.11 & 1.66 & 2.20 & 2.40 & 81.69 & 74.46 & 71.99 & 55.17 & 47.43 & 41.08 \\
FS $K=6$ + blur & 88.86 & 88.34 & 87.68 & 85.85 & 73.39 & 71.43 & 70.73 & 70.58 & 69.27 & 67.93 & 60.67 & 59.38 & 0.74 & 0.73 & 0.77 & 0.79 & 0.87 & 0.86 & 81.41 & 79.77 & 80.04 & 77.33 & 73.26 & 72.43 \\
FS $K=8$ & 56.49 & 38.66 & 24.90 & 6.68 & 6.42 & 4.85 & 58.64 & 49.99 & 41.78 & 31.01 & 28.87 & 29.13 & 0.98 & 1.09 & 1.52 & 2.35 & 2.90 & 2.91 & 80.59 & 62.58 & 57.13 & 31.76 & 27.02 & 22.65 \\
FS $K=8$ + blur & 88.47 & 88.20 & 87.42 & 85.19 & 72.08 & 69.46 & 70.37 & 69.87 & 69.04 & 67.32 & 59.41 & 58.04 & 0.74 & 0.74 & 0.78 & 0.80 & 0.89 & 0.88 & 81.20 & 79.41 & 80.72 & 76.60 & 74.30 & 72.24 \\
FS $K=10$ & 41.68 & 21.76 & 10.88 & 1.70 & 2.10 & 3.67 & 51.85 & 41.58 & 34.04 & 25.24 & 26.35 & 27.39 & 1.18 & 1.35 & 1.95 & 3.05 & 3.29 & 3.14 & 76.42 & 54.88 & 39.08 & 18.34 & 10.50 & 11.51 \\
FS $K=10$ + blur & 88.47 & 87.81 & 86.63 & 84.80 & 70.64 & 69.07 & 70.03 & 69.59 & 68.59 & 67.11 & 58.71 & 57.56 & 0.74 & 0.74 & 0.78 & 0.81 & 0.90 & 0.89 & 81.47 & 80.03 & 80.58 & 78.08 & 75.94 & 72.64 \\
FS $K=15$ & 20.45 & 4.98 & 1.31 & 0.26 & 0.79 & 1.97 & 38.25 & 23.88 & 23.00 & 18.67 & 23.30 & 25.50 & 1.79 & 2.27 & 3.17 & 4.47 & 3.74 & 3.42 & 56.74 & 16.96 & 7.83 & 4.62 & 2.84 & 4.60 \\
FS $K=15$ + blur & 88.73 & 88.07 & 86.63 & 84.14 & 70.12 & 67.50 & 69.63 & 69.33 & 68.29 & 66.58 & 58.13 & 57.00 & 0.74 & 0.74 & 0.78 & 0.81 & 0.91 & 0.90 & 81.80 & 80.39 & 80.16 & 78.71 & 75.37 & 72.71 \\
FS $K=20$ & 7.86 & 2.10 & 0.52 & 0.00 & 0.66 & 1.44 & 28.47 & 18.23 & 17.61 & 17.22 & 22.41 & 24.65 & 2.56 & 3.14 & 4.37 & 4.97 & 3.90 & 3.51 & 28.26 & 3.65 & 3.87 & 2.74 & 1.60 & 3.27 \\
FS $K=20$ + blur & 88.34 & 87.42 & 86.37 & 83.75 & 70.38 & 67.50 & 69.56 & 69.14 & 68.14 & 66.22 & 57.81 & 56.77 & 0.74 & 0.74 & 0.78 & 0.81 & 0.92 & 0.91 & 81.88 & 80.26 & 80.11 & 78.55 & 74.80 & 72.65 \\
\bottomrule
\end{tabular}
}
\end{table*}

%% file: Tables/TableQualitativeExamples.tex
\hypertarget{hyper:tab:qualitative}{\vspace*{0pt}}
\newcommand{\imgw}{0.13\linewidth}  %

\setlength{\LTcapwidth}{0.87\linewidth} %
\begin{longtable}{lcccccc}

\caption{Qualitative effect of Floyd--Steinberg quantization level $K$ on six PascalVOC images under clean and adversarial (SIA, $\epsilon_\infty{=}3/255$) conditions, using DINOv2 ViT-S/14. Each block displays the dithered input, classification prediction (green~$=$~correct, red~$=$~incorrect), and segmentation mask. The defense transitions from destructive ($K{=}2$) through effective ($K{=}3$) to transparent ($K{=}20$); at $K{=}6$ a transitional regime emerges where the adversarial target begins to dominate the softmax while the true class retains residual probability.}
\label{tab:qualitative_examples}\\

\toprule
 & No FS & $K{=}2$ & $K{=}3$ & $K{=}6$ & $K{=}9$ & $K{=}20$ \\ \midrule
\endfirsthead

\multicolumn{7}{c}{\small \tablename\ \thetable{} -- \textit{continued from previous page}}\\[4pt]
\toprule
 & No FS & $K{=}2$ & $K{=}3$ & $K{=}6$ & $K{=}9$ & $K{=}20$ \\ \midrule
\endhead

\midrule \multicolumn{7}{r}{\small\textit{Continued on next page}}\\
\endfoot

\bottomrule
\endlastfoot

\multicolumn{7}{l}{\hypertarget{hyper:qual:a}{}\textbf{(a)} Ground truth: \textit{car}} \\*[2pt]
\multirow{3}{*}[-42pt]{\rotatebox{90}{\textbf{Clean}}}
 & \adjustbox{valign=c}{\includegraphics[width=\imgw]{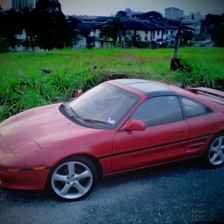}}
 & \adjustbox{valign=c}{\includegraphics[width=\imgw]{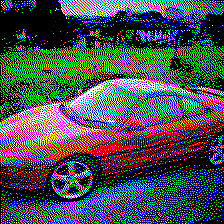}}
 & \adjustbox{valign=c}{\includegraphics[width=\imgw]{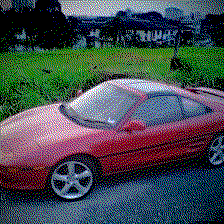}}
 & \adjustbox{valign=c}{\includegraphics[width=\imgw]{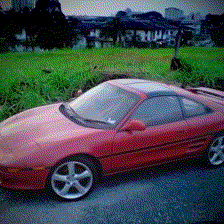}}
 & \adjustbox{valign=c}{\includegraphics[width=\imgw]{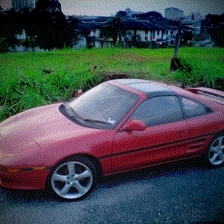}}
 & \adjustbox{valign=c}{\includegraphics[width=\imgw]{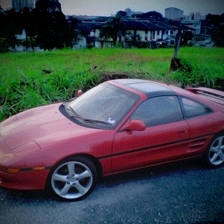}} \\*[2pt]
 & \parbox[c][2em][c]{\imgw}{\centering \scriptsize \textcolor{green}{car (99.9\%)}}
 & \parbox[c][2em][c]{\imgw}{\centering \scriptsize \textcolor{red}{pottedplant (45.9\%)} \\ car (6.7\%)}
 & \parbox[c][2em][c]{\imgw}{\centering \scriptsize \textcolor{green}{car (100.0\%)}}
 & \parbox[c][2em][c]{\imgw}{\centering \scriptsize \textcolor{green}{car (99.9\%)}}
 & \parbox[c][2em][c]{\imgw}{\centering \scriptsize \textcolor{green}{car (99.8\%)}}
 & \parbox[c][2em][c]{\imgw}{\centering \scriptsize \textcolor{green}{car (99.9\%)}} \\*[2pt]
 & \adjustbox{valign=c}{\includegraphics[width=\imgw]{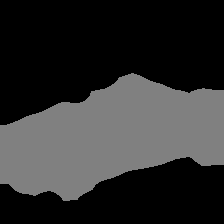}}
 & \adjustbox{valign=c}{\includegraphics[width=\imgw]{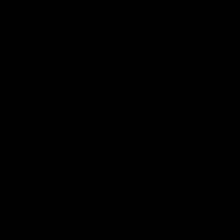}}
 & \adjustbox{valign=c}{\includegraphics[width=\imgw]{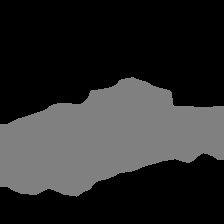}}
 & \adjustbox{valign=c}{\includegraphics[width=\imgw]{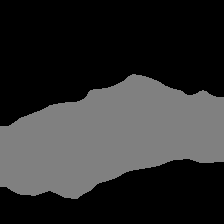}}
 & \adjustbox{valign=c}{\includegraphics[width=\imgw]{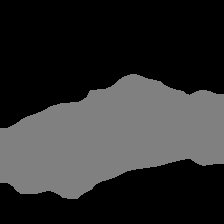}}
 & \adjustbox{valign=c}{\includegraphics[width=\imgw]{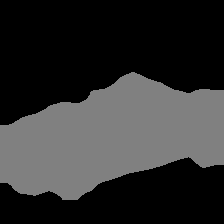}} \\*
\addlinespace[2pt]
\cdashline{1-7}
\addlinespace[2pt]
\multirow{3}{*}[-36pt]{\rotatebox{90}{\textbf{Adversarial}}}
 & \adjustbox{valign=c}{\includegraphics[width=\imgw]{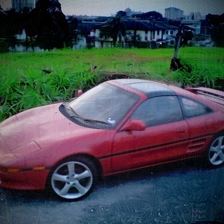}}
 & \adjustbox{valign=c}{\includegraphics[width=\imgw]{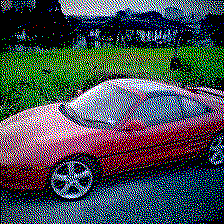}}
 & \adjustbox{valign=c}{\includegraphics[width=\imgw]{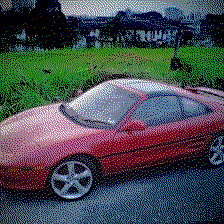}}
 & \adjustbox{valign=c}{\includegraphics[width=\imgw]{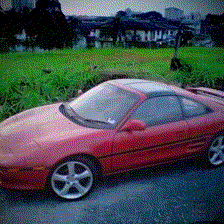}}
 & \adjustbox{valign=c}{\includegraphics[width=\imgw]{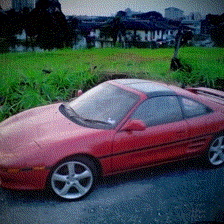}}
 & \adjustbox{valign=c}{\includegraphics[width=\imgw]{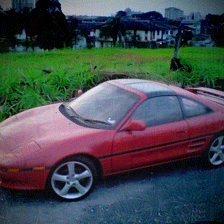}} \\*[2pt]
 & \parbox[c][2em][c]{\imgw}{\centering \scriptsize \textcolor{red}{horse (94.6\%)} \\ car (0.0\%)}
 & \parbox[c][2em][c]{\imgw}{\centering \scriptsize \textcolor{red}{cow (31.9\%)} \\ car (5.7\%)}
 & \parbox[c][2em][c]{\imgw}{\centering \scriptsize \textcolor{green}{car (99.8\%)}}
 & \parbox[c][2em][c]{\imgw}{\centering \scriptsize \textcolor{red}{motorbike (72.0\%)} \\ car (8.2\%)}
 & \parbox[c][2em][c]{\imgw}{\centering \scriptsize \textcolor{red}{horse (62.8\%)} \\ car (0.0\%)}
 & \parbox[c][2em][c]{\imgw}{\centering \scriptsize \textcolor{red}{horse (85.3\%)} \\ car (0.0\%)} \\*[2pt]
 & \adjustbox{valign=c}{\includegraphics[width=\imgw]{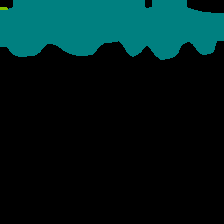}}
 & \adjustbox{valign=c}{\includegraphics[width=\imgw]{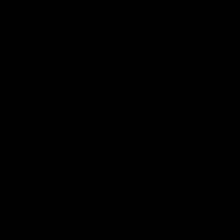}}
 & \adjustbox{valign=c}{\includegraphics[width=\imgw]{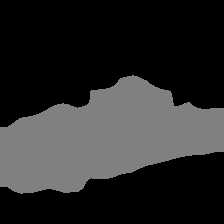}}
 & \adjustbox{valign=c}{\includegraphics[width=\imgw]{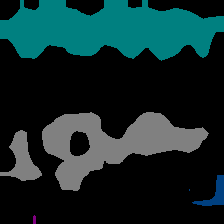}}
 & \adjustbox{valign=c}{\includegraphics[width=\imgw]{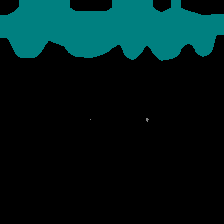}}
 & \adjustbox{valign=c}{\includegraphics[width=\imgw]{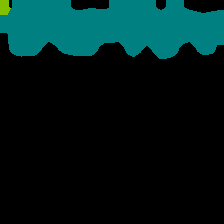}} \\
\pagebreak

\multicolumn{7}{l}{\hypertarget{hyper:qual:b}{}\textbf{(b)} Ground truth: \textit{dog}} \\*[2pt]
\multirow{3}{*}[-42pt]{\rotatebox{90}{\textbf{Clean}}}
 & \adjustbox{valign=c}{\includegraphics[width=\imgw]{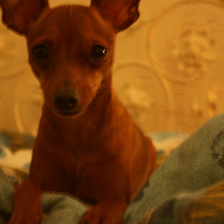}}
 & \adjustbox{valign=c}{\includegraphics[width=\imgw]{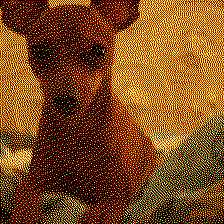}}
 & \adjustbox{valign=c}{\includegraphics[width=\imgw]{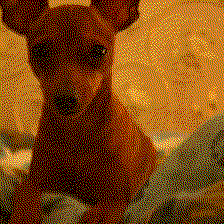}}
 & \adjustbox{valign=c}{\includegraphics[width=\imgw]{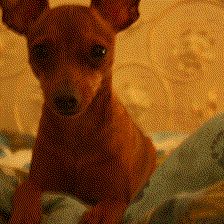}}
 & \adjustbox{valign=c}{\includegraphics[width=\imgw]{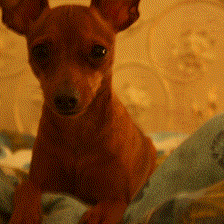}}
 & \adjustbox{valign=c}{\includegraphics[width=\imgw]{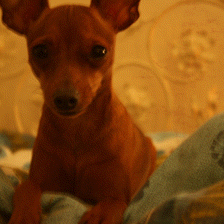}} \\*[2pt]
 & \parbox[c][2em][c]{\imgw}{\centering \scriptsize \textcolor{green}{dog (99.9\%)}}
 & \parbox[c][2em][c]{\imgw}{\centering \scriptsize \textcolor{red}{bottle (26.7\%)} \\ dog (1.8\%)}
 & \parbox[c][2em][c]{\imgw}{\centering \scriptsize \textcolor{green}{dog (99.9\%)}}
 & \parbox[c][2em][c]{\imgw}{\centering \scriptsize \textcolor{green}{dog (99.8\%)}}
 & \parbox[c][2em][c]{\imgw}{\centering \scriptsize \textcolor{green}{dog (99.9\%)}}
 & \parbox[c][2em][c]{\imgw}{\centering \scriptsize \textcolor{green}{dog (99.9\%)}} \\*[2pt]
 & \adjustbox{valign=c}{\includegraphics[width=\imgw]{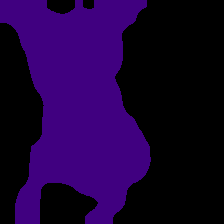}}
 & \adjustbox{valign=c}{\includegraphics[width=\imgw]{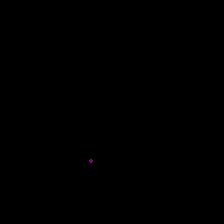}}
 & \adjustbox{valign=c}{\includegraphics[width=\imgw]{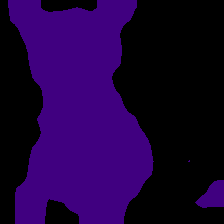}}
 & \adjustbox{valign=c}{\includegraphics[width=\imgw]{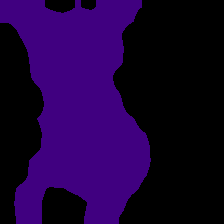}}
 & \adjustbox{valign=c}{\includegraphics[width=\imgw]{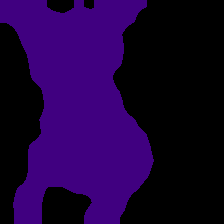}}
 & \adjustbox{valign=c}{\includegraphics[width=\imgw]{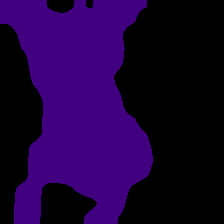}} \\*
\addlinespace[2pt]
\cdashline{1-7}
\addlinespace[2pt]
\multirow{3}{*}[-36pt]{\rotatebox{90}{\textbf{Adversarial}}}
 & \adjustbox{valign=c}{\includegraphics[width=\imgw]{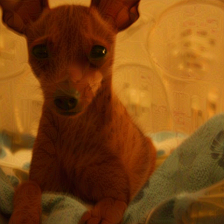}}
 & \adjustbox{valign=c}{\includegraphics[width=\imgw]{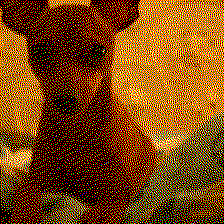}}
 & \adjustbox{valign=c}{\includegraphics[width=\imgw]{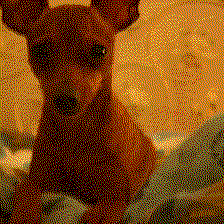}}
 & \adjustbox{valign=c}{\includegraphics[width=\imgw]{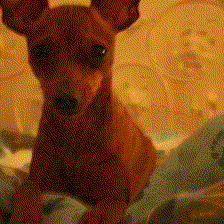}}
 & \adjustbox{valign=c}{\includegraphics[width=\imgw]{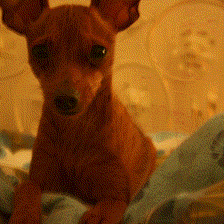}}
 & \adjustbox{valign=c}{\includegraphics[width=\imgw]{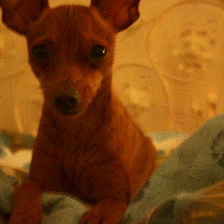}} \\*[2pt]
 & \parbox[c][2em][c]{\imgw}{\centering \scriptsize \textcolor{red}{cat (100.0\%)} \\ dog (0.0\%)}
 & \parbox[c][2em][c]{\imgw}{\centering \scriptsize \textcolor{red}{bottle (32.5\%)} \\ dog (0.8\%)}
 & \parbox[c][2em][c]{\imgw}{\centering \scriptsize \textcolor{green}{dog (99.8\%)}}
 & \parbox[c][2em][c]{\imgw}{\centering \scriptsize \textcolor{red}{cat (99.3\%)} \\ dog (0.6\%)}
 & \parbox[c][2em][c]{\imgw}{\centering \scriptsize \textcolor{red}{cat (100.0\%)} \\ dog (0.0\%)}
 & \parbox[c][2em][c]{\imgw}{\centering \scriptsize \textcolor{red}{cat (100.0\%)} \\ dog (0.0\%)} \\*[2pt]
 & \adjustbox{valign=c}{\includegraphics[width=\imgw]{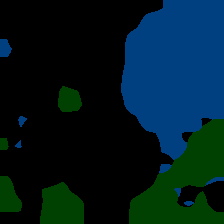}}
 & \adjustbox{valign=c}{\includegraphics[width=\imgw]{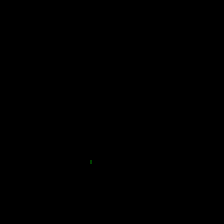}}
 & \adjustbox{valign=c}{\includegraphics[width=\imgw]{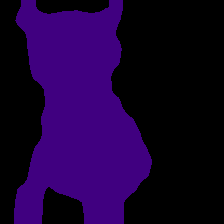}}
 & \adjustbox{valign=c}{\includegraphics[width=\imgw]{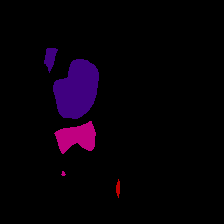}}
 & \adjustbox{valign=c}{\includegraphics[width=\imgw]{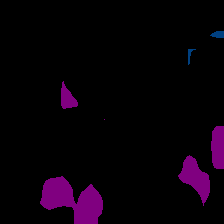}}
 & \adjustbox{valign=c}{\includegraphics[width=\imgw]{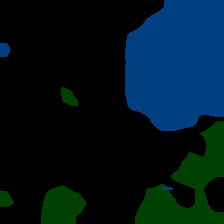}} \\
\pagebreak

\multicolumn{7}{l}{\hypertarget{hyper:qual:c}{}\textbf{(c)} Ground truth: \textit{person}} \\*[2pt]
\multirow{3}{*}[-42pt]{\rotatebox{90}{\textbf{Clean}}}
 & \adjustbox{valign=c}{\includegraphics[width=\imgw]{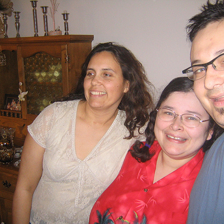}}
 & \adjustbox{valign=c}{\includegraphics[width=\imgw]{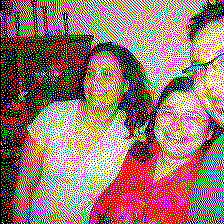}}
 & \adjustbox{valign=c}{\includegraphics[width=\imgw]{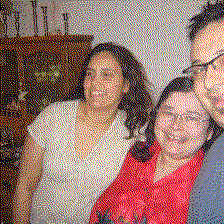}}
 & \adjustbox{valign=c}{\includegraphics[width=\imgw]{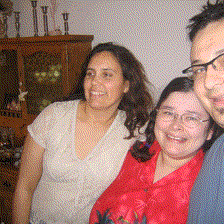}}
 & \adjustbox{valign=c}{\includegraphics[width=\imgw]{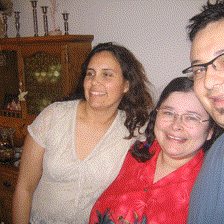}}
 & \adjustbox{valign=c}{\includegraphics[width=\imgw]{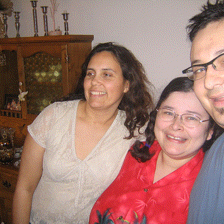}} \\*[2pt]
 & \parbox[c][2em][c]{\imgw}{\centering \scriptsize \textcolor{green}{person (98.4\%)}}
 & \parbox[c][2em][c]{\imgw}{\centering \scriptsize \textcolor{red}{pottedplant (29.7\%)} \\ person (19.6\%)}
 & \parbox[c][2em][c]{\imgw}{\centering \scriptsize \textcolor{green}{person (99.3\%)}}
 & \parbox[c][2em][c]{\imgw}{\centering \scriptsize \textcolor{green}{person (99.0\%)}}
 & \parbox[c][2em][c]{\imgw}{\centering \scriptsize \textcolor{green}{person (98.9\%)}}
 & \parbox[c][2em][c]{\imgw}{\centering \scriptsize \textcolor{green}{person (98.7\%)}} \\*[2pt]
 & \adjustbox{valign=c}{\includegraphics[width=\imgw]{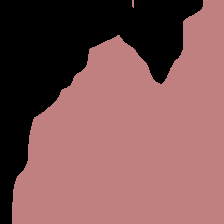}}
 & \adjustbox{valign=c}{\includegraphics[width=\imgw]{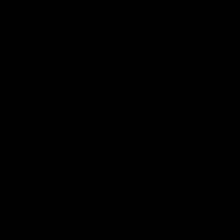}}
 & \adjustbox{valign=c}{\includegraphics[width=\imgw]{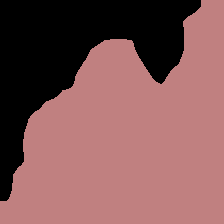}}
 & \adjustbox{valign=c}{\includegraphics[width=\imgw]{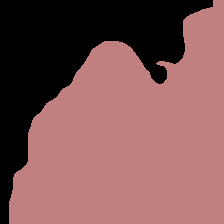}}
 & \adjustbox{valign=c}{\includegraphics[width=\imgw]{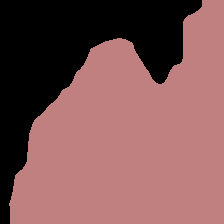}}
 & \adjustbox{valign=c}{\includegraphics[width=\imgw]{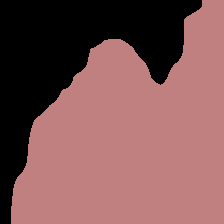}} \\*
\addlinespace[2pt]
\cdashline{1-7}
\addlinespace[2pt]
\multirow{3}{*}[-36pt]{\rotatebox{90}{\textbf{Adversarial}}}
 & \adjustbox{valign=c}{\includegraphics[width=\imgw]{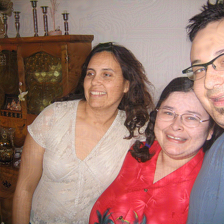}}
 & \adjustbox{valign=c}{\includegraphics[width=\imgw]{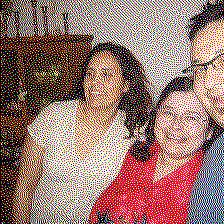}}
 & \adjustbox{valign=c}{\includegraphics[width=\imgw]{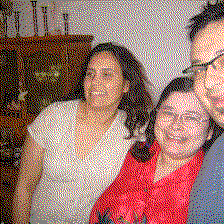}}
 & \adjustbox{valign=c}{\includegraphics[width=\imgw]{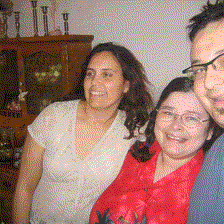}}
 & \adjustbox{valign=c}{\includegraphics[width=\imgw]{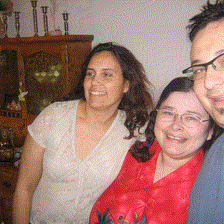}}
 & \adjustbox{valign=c}{\includegraphics[width=\imgw]{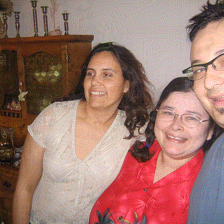}} \\*[2pt]
 & \parbox[c][2em][c]{\imgw}{\centering \scriptsize \textcolor{red}{bottle (100.0\%)} \\ person (0.0\%)}
 & \parbox[c][2em][c]{\imgw}{\centering \scriptsize \textcolor{red}{pottedplant (26.2\%)} \\ person (20.4\%)}
 & \parbox[c][2em][c]{\imgw}{\centering \scriptsize \textcolor{green}{person (98.8\%)}}
 & \parbox[c][2em][c]{\imgw}{\centering \scriptsize \textcolor{red}{bottle (68.6\%)} \\ person (28.3\%)}
 & \parbox[c][2em][c]{\imgw}{\centering \scriptsize \textcolor{red}{bottle (99.9\%)} \\ person (0.0\%)}
 & \parbox[c][2em][c]{\imgw}{\centering \scriptsize \textcolor{red}{bottle (100.0\%)} \\ person (0.0\%)} \\*[2pt]
 & \adjustbox{valign=c}{\includegraphics[width=\imgw]{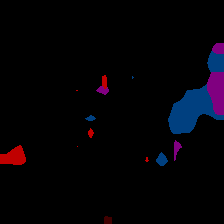}}
 & \adjustbox{valign=c}{\includegraphics[width=\imgw]{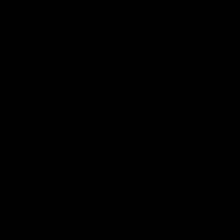}}
 & \adjustbox{valign=c}{\includegraphics[width=\imgw]{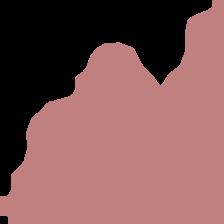}}
 & \adjustbox{valign=c}{\includegraphics[width=\imgw]{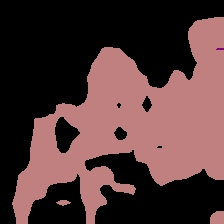}}
 & \adjustbox{valign=c}{\includegraphics[width=\imgw]{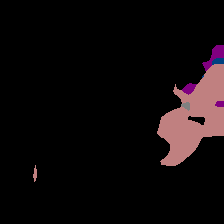}}
 & \adjustbox{valign=c}{\includegraphics[width=\imgw]{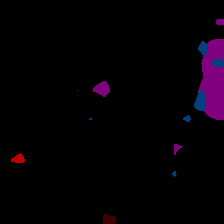}} \\
\pagebreak

\multicolumn{7}{l}{\hypertarget{hyper:qual:d}{}\textbf{(d)} Ground truth: \textit{cow}} \\*[2pt]
\multirow{3}{*}[-42pt]{\rotatebox{90}{\textbf{Clean}}}
 & \adjustbox{valign=c}{\includegraphics[width=\imgw]{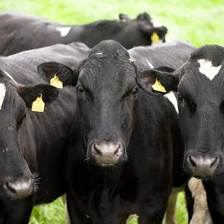}}
 & \adjustbox{valign=c}{\includegraphics[width=\imgw]{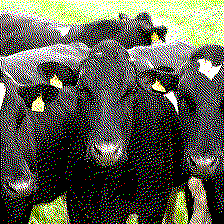}}
 & \adjustbox{valign=c}{\includegraphics[width=\imgw]{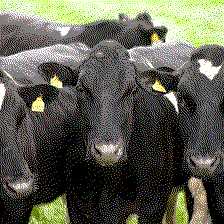}}
 & \adjustbox{valign=c}{\includegraphics[width=\imgw]{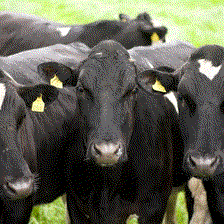}}
 & \adjustbox{valign=c}{\includegraphics[width=\imgw]{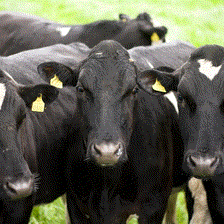}}
 & \adjustbox{valign=c}{\includegraphics[width=\imgw]{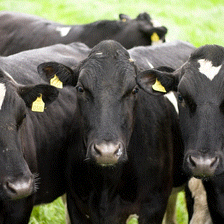}} \\*[2pt]
 & \parbox[c][2em][c]{\imgw}{\centering \scriptsize \textcolor{green}{cow (100.0\%)}}
 & \parbox[c][2em][c]{\imgw}{\centering \scriptsize \textcolor{red}{sheep (69.0\%)} \\ cow (24.0\%)}
 & \parbox[c][2em][c]{\imgw}{\centering \scriptsize \textcolor{green}{cow (99.8\%)}}
 & \parbox[c][2em][c]{\imgw}{\centering \scriptsize \textcolor{green}{cow (100.0\%)}}
 & \parbox[c][2em][c]{\imgw}{\centering \scriptsize \textcolor{green}{cow (100.0\%)}}
 & \parbox[c][2em][c]{\imgw}{\centering \scriptsize \textcolor{green}{cow (100.0\%)}} \\*[2pt]
 & \adjustbox{valign=c}{\includegraphics[width=\imgw]{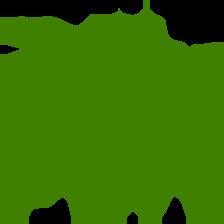}}
 & \adjustbox{valign=c}{\includegraphics[width=\imgw]{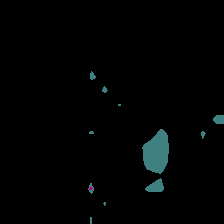}}
 & \adjustbox{valign=c}{\includegraphics[width=\imgw]{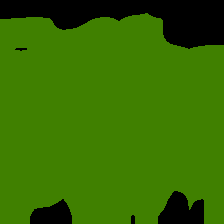}}
 & \adjustbox{valign=c}{\includegraphics[width=\imgw]{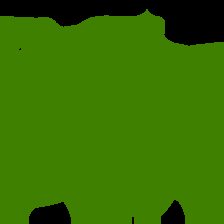}}
 & \adjustbox{valign=c}{\includegraphics[width=\imgw]{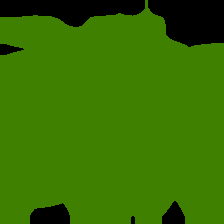}}
 & \adjustbox{valign=c}{\includegraphics[width=\imgw]{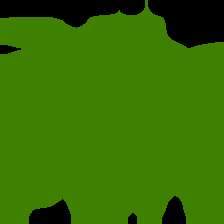}} \\*
\addlinespace[2pt]
\cdashline{1-7}
\addlinespace[2pt]
\multirow{3}{*}[-36pt]{\rotatebox{90}{\textbf{Adversarial}}}
 & \adjustbox{valign=c}{\includegraphics[width=\imgw]{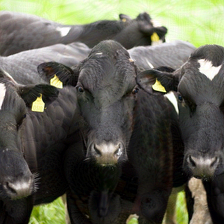}}
 & \adjustbox{valign=c}{\includegraphics[width=\imgw]{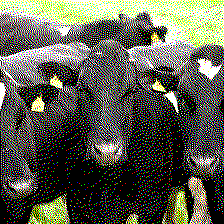}}
 & \adjustbox{valign=c}{\includegraphics[width=\imgw]{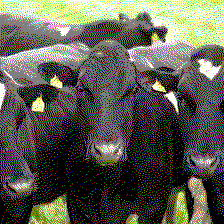}}
 & \adjustbox{valign=c}{\includegraphics[width=\imgw]{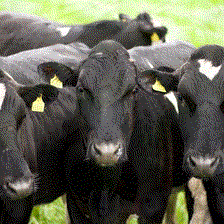}}
 & \adjustbox{valign=c}{\includegraphics[width=\imgw]{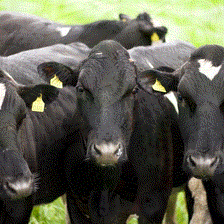}}
 & \adjustbox{valign=c}{\includegraphics[width=\imgw]{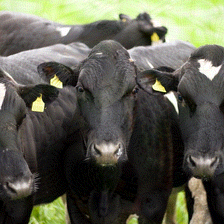}} \\*[2pt]
 & \parbox[c][2em][c]{\imgw}{\centering \scriptsize \textcolor{red}{bird (100.0\%)} \\ cow (0.0\%)}
 & \parbox[c][2em][c]{\imgw}{\centering \scriptsize \textcolor{red}{sheep (74.4\%)} \\ cow (1.3\%)}
 & \parbox[c][2em][c]{\imgw}{\centering \scriptsize \textcolor{green}{cow (83.1\%)}}
 & \parbox[c][2em][c]{\imgw}{\centering \scriptsize \textcolor{red}{sheep (59.9\%)} \\ cow (18.5\%)}
 & \parbox[c][2em][c]{\imgw}{\centering \scriptsize \textcolor{red}{bird (100.0\%)} \\ cow (0.0\%)}
 & \parbox[c][2em][c]{\imgw}{\centering \scriptsize \textcolor{red}{bird (100.0\%)} \\ cow (0.0\%)} \\*[2pt]
 & \adjustbox{valign=c}{\includegraphics[width=\imgw]{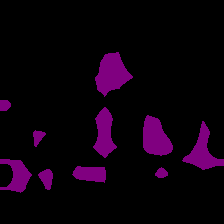}}
 & \adjustbox{valign=c}{\includegraphics[width=\imgw]{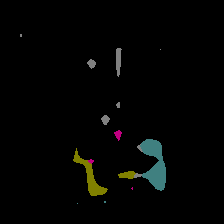}}
 & \adjustbox{valign=c}{\includegraphics[width=\imgw]{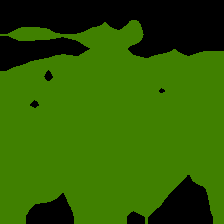}}
 & \adjustbox{valign=c}{\includegraphics[width=\imgw]{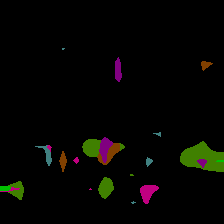}}
 & \adjustbox{valign=c}{\includegraphics[width=\imgw]{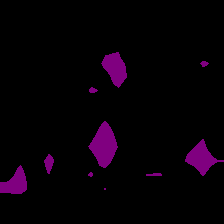}}
 & \adjustbox{valign=c}{\includegraphics[width=\imgw]{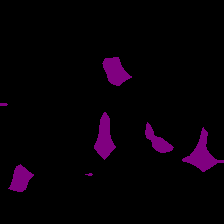}} \\
\pagebreak

\multicolumn{7}{l}{\hypertarget{hyper:qual:e}{}\textbf{(e)} Ground truth: \textit{cat}} \\*[2pt]
\multirow{3}{*}[-42pt]{\rotatebox{90}{\textbf{Clean}}}
 & \adjustbox{valign=c}{\includegraphics[width=\imgw]{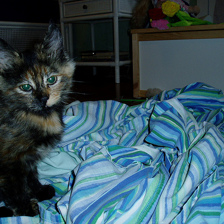}}
 & \adjustbox{valign=c}{\includegraphics[width=\imgw]{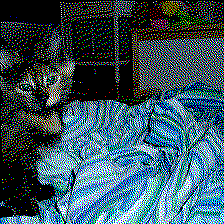}}
 & \adjustbox{valign=c}{\includegraphics[width=\imgw]{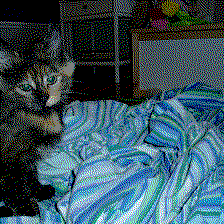}}
 & \adjustbox{valign=c}{\includegraphics[width=\imgw]{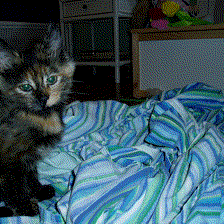}}
 & \adjustbox{valign=c}{\includegraphics[width=\imgw]{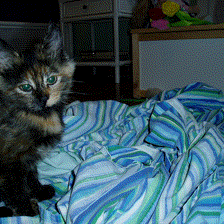}}
 & \adjustbox{valign=c}{\includegraphics[width=\imgw]{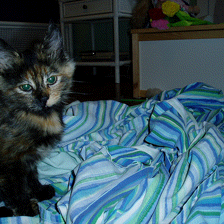}} \\*[2pt]
 & \parbox[c][2em][c]{\imgw}{\centering \scriptsize \textcolor{green}{cat (100.0\%)}}
 & \parbox[c][2em][c]{\imgw}{\centering \scriptsize \textcolor{red}{pottedplant (58.8\%)} \\ cat (28.0\%)}
 & \parbox[c][2em][c]{\imgw}{\centering \scriptsize \textcolor{green}{cat (100.0\%)}}
 & \parbox[c][2em][c]{\imgw}{\centering \scriptsize \textcolor{green}{cat (100.0\%)}}
 & \parbox[c][2em][c]{\imgw}{\centering \scriptsize \textcolor{green}{cat (100.0\%)}}
 & \parbox[c][2em][c]{\imgw}{\centering \scriptsize \textcolor{green}{cat (100.0\%)}} \\*[2pt]
 & \adjustbox{valign=c}{\includegraphics[width=\imgw]{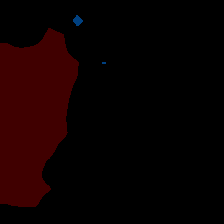}}
 & \adjustbox{valign=c}{\includegraphics[width=\imgw]{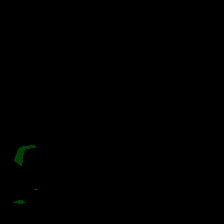}}
 & \adjustbox{valign=c}{\includegraphics[width=\imgw]{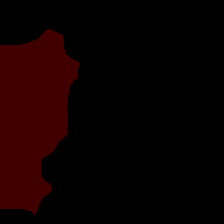}}
 & \adjustbox{valign=c}{\includegraphics[width=\imgw]{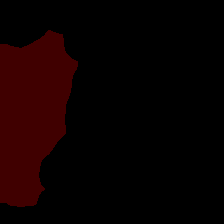}}
 & \adjustbox{valign=c}{\includegraphics[width=\imgw]{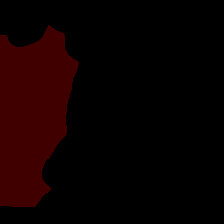}}
 & \adjustbox{valign=c}{\includegraphics[width=\imgw]{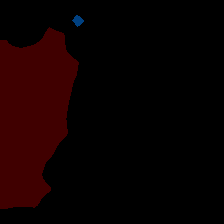}} \\*
\addlinespace[2pt]
\cdashline{1-7}
\addlinespace[2pt]
\multirow{3}{*}[-36pt]{\rotatebox{90}{\textbf{Adversarial}}}
 & \adjustbox{valign=c}{\includegraphics[width=\imgw]{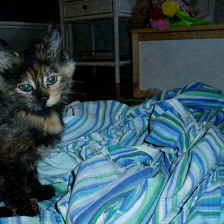}}
 & \adjustbox{valign=c}{\includegraphics[width=\imgw]{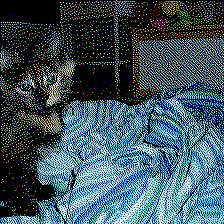}}
 & \adjustbox{valign=c}{\includegraphics[width=\imgw]{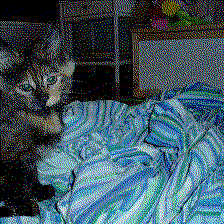}}
 & \adjustbox{valign=c}{\includegraphics[width=\imgw]{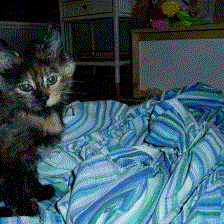}}
 & \adjustbox{valign=c}{\includegraphics[width=\imgw]{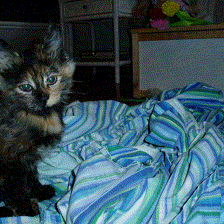}}
 & \adjustbox{valign=c}{\includegraphics[width=\imgw]{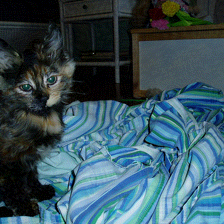}} \\*[2pt]
 & \parbox[c][2em][c]{\imgw}{\centering \scriptsize \textcolor{red}{dog (100.0\%)} \\ cat (0.0\%)}
 & \parbox[c][2em][c]{\imgw}{\centering \scriptsize \textcolor{red}{pottedplant (57.0\%)} \\ cat (29.0\%)}
 & \parbox[c][2em][c]{\imgw}{\centering \scriptsize \textcolor{green}{cat (99.7\%)}}
 & \parbox[c][2em][c]{\imgw}{\centering \scriptsize \textcolor{red}{dog (98.4\%)} \\ cat (1.3\%)}
 & \parbox[c][2em][c]{\imgw}{\centering \scriptsize \textcolor{red}{dog (100.0\%)} \\ cat (0.0\%)}
 & \parbox[c][2em][c]{\imgw}{\centering \scriptsize \textcolor{red}{dog (100.0\%)} \\ cat (0.0\%)} \\*[2pt]
 & \adjustbox{valign=c}{\includegraphics[width=\imgw]{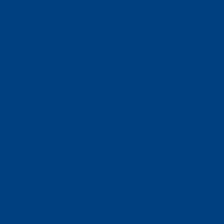}}
 & \adjustbox{valign=c}{\includegraphics[width=\imgw]{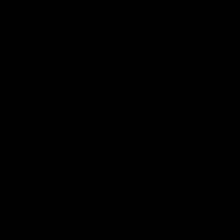}}
 & \adjustbox{valign=c}{\includegraphics[width=\imgw]{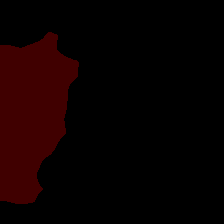}}
 & \adjustbox{valign=c}{\includegraphics[width=\imgw]{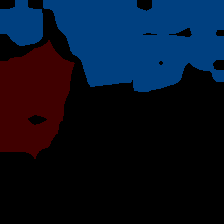}}
 & \adjustbox{valign=c}{\includegraphics[width=\imgw]{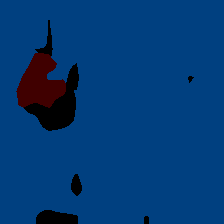}}
 & \adjustbox{valign=c}{\includegraphics[width=\imgw]{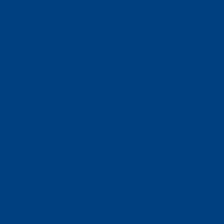}} \\
\pagebreak

\multicolumn{7}{l}{\hypertarget{hyper:qual:f}{}\textbf{(f)} Ground truth: \textit{aeroplane}} \\*[2pt]
\multirow{3}{*}[-42pt]{\rotatebox{90}{\textbf{Clean}}}
 & \adjustbox{valign=c}{\includegraphics[width=\imgw]{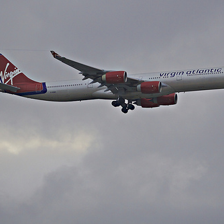}}
 & \adjustbox{valign=c}{\includegraphics[width=\imgw]{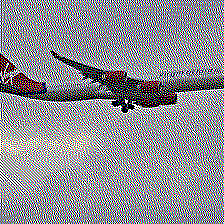}}
 & \adjustbox{valign=c}{\includegraphics[width=\imgw]{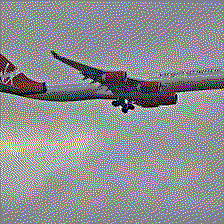}}
 & \adjustbox{valign=c}{\includegraphics[width=\imgw]{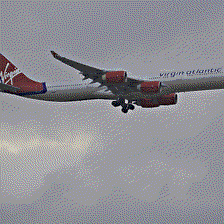}}
 & \adjustbox{valign=c}{\includegraphics[width=\imgw]{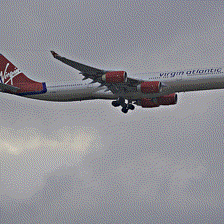}}
 & \adjustbox{valign=c}{\includegraphics[width=\imgw]{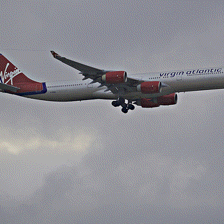}} \\*[2pt]
 & \parbox[c][2em][c]{\imgw}{\centering \scriptsize \textcolor{green}{aeroplane (99.9\%)}}
 & \parbox[c][2em][c]{\imgw}{\centering \scriptsize \textcolor{red}{cat (41.9\%)} \\ aeroplane (0.6\%)}
 & \parbox[c][2em][c]{\imgw}{\centering \scriptsize \textcolor{green}{aeroplane (100.0\%)}}
 & \parbox[c][2em][c]{\imgw}{\centering \scriptsize \textcolor{green}{aeroplane (100.0\%)}}
 & \parbox[c][2em][c]{\imgw}{\centering \scriptsize \textcolor{green}{aeroplane (100.0\%)}}
 & \parbox[c][2em][c]{\imgw}{\centering \scriptsize \textcolor{green}{aeroplane (99.9\%)}} \\*[2pt]
 & \adjustbox{valign=c}{\includegraphics[width=\imgw]{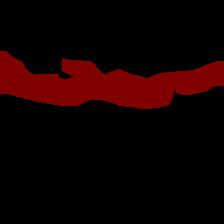}}
 & \adjustbox{valign=c}{\includegraphics[width=\imgw]{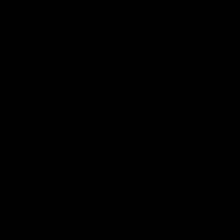}}
 & \adjustbox{valign=c}{\includegraphics[width=\imgw]{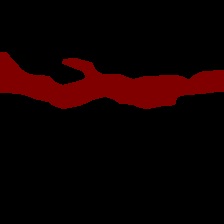}}
 & \adjustbox{valign=c}{\includegraphics[width=\imgw]{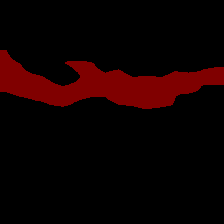}}
 & \adjustbox{valign=c}{\includegraphics[width=\imgw]{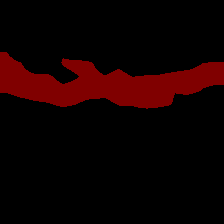}}
 & \adjustbox{valign=c}{\includegraphics[width=\imgw]{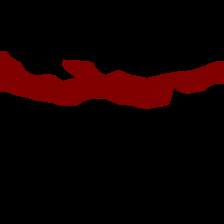}} \\*
\addlinespace[2pt]
\cdashline{1-7}
\addlinespace[2pt]
\multirow{3}{*}[-36pt]{\rotatebox{90}{\textbf{Adversarial}}}
 & \adjustbox{valign=c}{\includegraphics[width=\imgw]{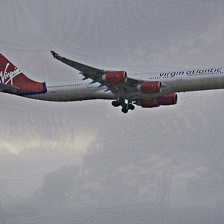}}
 & \adjustbox{valign=c}{\includegraphics[width=\imgw]{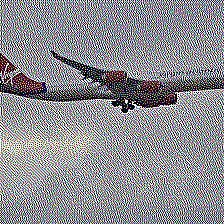}}
 & \adjustbox{valign=c}{\includegraphics[width=\imgw]{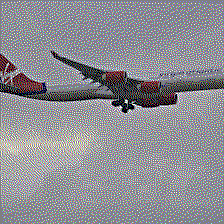}}
 & \adjustbox{valign=c}{\includegraphics[width=\imgw]{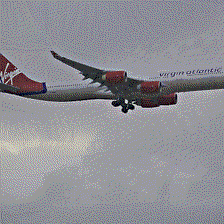}}
 & \adjustbox{valign=c}{\includegraphics[width=\imgw]{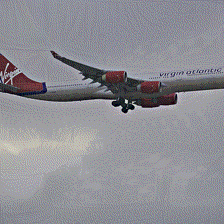}}
 & \adjustbox{valign=c}{\includegraphics[width=\imgw]{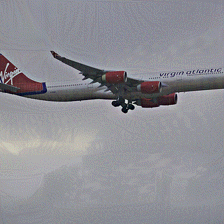}} \\*[2pt]
 & \parbox[c][2em][c]{\imgw}{\centering \scriptsize \textcolor{red}{bottle (79.8\%)} \\ aeroplane (0.0\%)}
 & \parbox[c][2em][c]{\imgw}{\centering \scriptsize \textcolor{red}{cat (51.6\%)} \\ aeroplane (0.3\%)}
 & \parbox[c][2em][c]{\imgw}{\centering \scriptsize \textcolor{green}{aeroplane (100.0\%)}}
 & \parbox[c][2em][c]{\imgw}{\centering \scriptsize \textcolor{green}{aeroplane (98.4\%)}}
 & \parbox[c][2em][c]{\imgw}{\centering \scriptsize \textcolor{green}{aeroplane (91.8\%)}}
 & \parbox[c][2em][c]{\imgw}{\centering \scriptsize \textcolor{red}{bottle (76.1\%)} \\ aeroplane (0.0\%)} \\*[2pt]
 & \adjustbox{valign=c}{\includegraphics[width=\imgw]{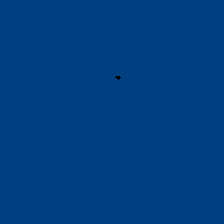}}
 & \adjustbox{valign=c}{\includegraphics[width=\imgw]{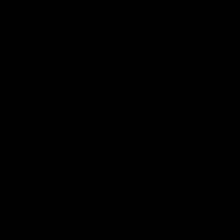}}
 & \adjustbox{valign=c}{\includegraphics[width=\imgw]{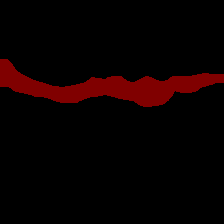}}
 & \adjustbox{valign=c}{\includegraphics[width=\imgw]{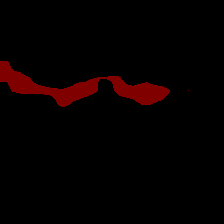}}
 & \adjustbox{valign=c}{\includegraphics[width=\imgw]{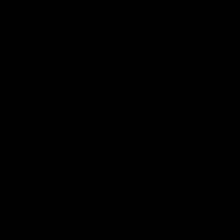}}
 & \adjustbox{valign=c}{\includegraphics[width=\imgw]{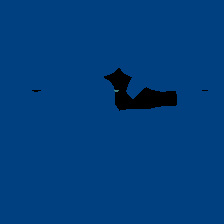}} \\

\end{longtable}